%% file: main.tex
\documentclass[10pt,twocolumn,letterpaper]{article}

\usepackage[pagenumbers]{cvpr}

\input{preamble}
\definecolor{cvprblue}{rgb}{0.21,0.49,0.74}
\usepackage[pagebackref,breaklinks,colorlinks,allcolors=cvprblue]{hyperref}
\hypersetup{
  pdftitle={NERVFace: Native-topology Face Reconstruction via Surface Correspondence},
  pdfauthor={Aocheng Huang, Chenchu Rong, Y.M. Wang, Guoqing Deng, Peng Xinghui, Jiawei Lu, Tianshu Wang, Jing Li, Huan Fu}
}

\def\confName{CVPR}
\def\confYear{2027}

\title{NERVFace: Native-topology Face Reconstruction\\
via Surface Correspondence}

\author{
Aocheng Huang$^{1,2}$\quad
Chenchu Rong$^{1}$\quad
Y.M.\ Wang$^{1,\dagger}$\quad
Guoqing Deng$^{2}$\quad
Peng Xinghui$^{1}$\\
Jiawei Lu$^{1}$\quad
Tianshu Wang$^{1}$\quad
Jing Li$^{1}$\quad
Huan Fu$^{3,\ddagger}$\\
{\small $^{1}$Alibaba Group ({\tt alibaba-inc.com})\quad
$^{2}$Nanjing University ({\tt nju.edu.cn})\quad
$^{3}$University of Sydney ({\tt usyd.edu.au})}\\
{\small $^{\dagger\ddagger}$ Project Lead
 \qquad Project page: \url{https://hacetate.github.io/NERVFace/}}
}

\begin{document}
\maketitle
\input{fig/teaser}
\input{sec/0_abstract}
\input{sec/1_intro}
\input{sec/2_related}
\input{sec/3_method}
\input{sec/4_experiments}
\input{sec/5_conclusion}
\input{sec/6_llm_statement}
{
    \small
    \bibliographystyle{ieeenat_fullname}
    \bibliography{main}
}
\clearpage
\appendix
\setcounter{figure}{0}
\setcounter{table}{1}
\setcounter{equation}{0}
\renewcommand{\thefigure}{S\arabic{figure}}
\renewcommand{\thetable}{S\arabic{table}}
\renewcommand{\theequation}{S\arabic{equation}}
\renewcommand{\theHfigure}{supp.\arabic{figure}}
\renewcommand{\theHtable}{supp.\arabic{table}}
\renewcommand{\theHequation}{supp.\arabic{equation}}
\input{sec/X_suppl}

\end{document}

%% file: preamble.tex
\usepackage{cuted}    %
\usepackage{currfile} %
\usepackage{caption}  %

\usepackage{multirow}
\usepackage{makecell}
\usepackage{bm}
\usepackage{pifont}
\usepackage{microtype}

\DeclareMathOperator{\LN}{LN}
\DeclareMathOperator{\RMS}{RMSNorm}

\DeclareMathOperator{\softmax}{softmax}

\newcommand{\method}{NERVFace\xspace}

\newcommand{\best}[1]{\textbf{#1}}
\newcommand{\second}[1]{\underline{#1}}
\newcommand{\suppref}[1]{Appendix~\ref{#1}}

%% file: fig/teaser.tex
\begin{strip}
\centering
\includegraphics[width=\linewidth]{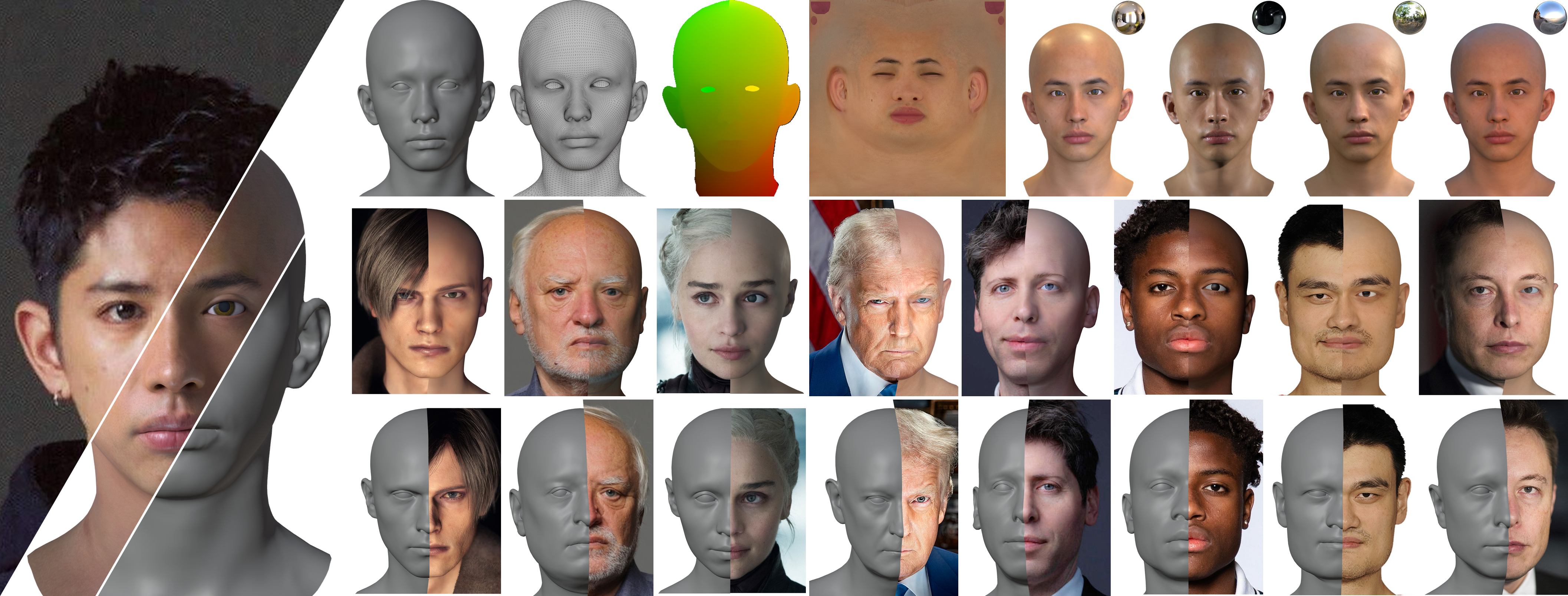}
\captionsetup{hypcap=false}
\captionof{figure}{\method reconstructs complete heads with fine facial detail in a shared topology and UV layout, with an albedo that relights consistently across environments. Left: input photograph, our textured render, and our mesh. Top right: mesh, wireframe, predicted UV coordinates, albedo atlas, and renders under four HDR maps (insets). Middle and bottom right: further in-the-wild portraits, each split between the photograph and our textured render (middle) or our mesh (bottom).}
\captionsetup{hypcap=true}
\label{fig:teaser}%
\end{strip}

%% file: sec/0_abstract.tex
\begin{abstract}
Reconstructing a textured 3D head from a single portrait while preserving a predefined topology requires consistent correspondence between image observations, geometry, and appearance. We present NERVFace, a correspondence-aware generative framework that leverages a predefined template's mesh topology and UV layout to guide head geometry and texture generation. For geometry generation, we exploit the spatial guidance provided by canonical surface coordinates. To incorporate this guidance, we train a Canonical Correspondence Encoder (CCE) with dense surface-coordinate supervision. Its intermediate representations condition a rectified flow over template-cluster latents to generate a native-topology head. For texture synthesis, visible portrait colors are projected through the reconstructed mesh to form a partial UV atlas that guides generation with a UV-adapted image-editing model. A geometry-attention adapter further incorporates geometric context from the reconstructed surface, guided by the fixed correspondence between atlas regions and the template mesh. Together, these complementary forms of surface correspondence link image evidence to geometry and geometry to appearance. Experiments demonstrate accurate geometry reconstruction and high-fidelity texture generation, producing detailed and consistent textured 3D heads from single portraits.
\end{abstract}

%% file: sec/1_intro.tex
\section{Introduction}
\label{sec:intro}

Reconstructing a textured 3D head from a single portrait is a fundamental task in digital character creation. Beyond recovering plausible geometry and appearance, practical applications often require a predefined mesh topology and UV layout, as in MetaHuman~\cite{metahuman}, to support consistent character assets, animation, and editing. The goal is to produce both a head mesh and its corresponding texture, while preserving the relationships between image observations, geometry, and appearance. Surface correspondence provides a natural basis for connecting these elements.

Existing methods address different aspects of this task. Parametric models~\cite{deca,emoca,mica,smirk,hrn} reconstruct facial geometry within a constrained shape space, while dense correspondence methods~\cite{flowface,pixel3dmm} use surface predictions to guide model fitting. Generative approaches~\cite{trellis2,hunyuan3d21} produce detailed 3D shapes, and TOPOS~\cite{topos} generates heads with a predefined topology using pretrained visual features~\cite{dinov2} and UV texture generation. For appearance synthesis, UV texture methods~\cite{ffhquv,uvidm,freeuv} generate textures in the atlas domain, sometimes incorporating geometric information through UV-space representations. These methods demonstrate the potential of generative head reconstruction. We explore how the correspondence defined by a common template can guide both geometry generation and texture synthesis.

We present \method, a correspondence-aware generative framework for reconstructing complete textured 3D heads from single portraits. Our central idea is to leverage a predefined template's mesh topology and UV layout as a common structural reference throughout generation (\cref{fig:teaser}). Learned image-to-surface correspondence guides geometry generation, while the known correspondence between the mesh and UV layout provides geometric context for texture synthesis. These complementary mechanisms connect image observations, geometry, and appearance within a consistent representation.

For geometry generation, we exploit the spatial guidance provided by canonical surface coordinates, which associate image observations with regions of the template surface. We introduce a Canonical Correspondence Encoder (CCE) trained with dense surface-coordinate supervision and use its intermediate representations to condition a rectified flow over template-cluster latents. Rather than using predicted coordinates as fitting constraints, this allows correspondence guidance to inform generative head reconstruction. For texture synthesis, visible portrait colors are projected through the reconstructed mesh to form a partial UV atlas, which guides generation with a UV-adapted image-editing model. A geometry-attention adapter further incorporates context from the reconstructed surface, guided by the known correspondence between the UV atlas and template mesh. Texture generation thus draws on both observed appearance and reconstructed geometry.

We evaluate \method on synthetic and real datasets using the MetaHuman topology. Results demonstrate accurate head geometry and improved texture fidelity, with competitive performance against existing methods. Qualitative examples further show detailed and consistent textured heads reconstructed from single portraits.

Our main contributions are:
\begin{itemize}
    \item We present a correspondence-aware generative framework for reconstructing complete textured 3D heads, using a predefined topology and UV layout to connect geometry generation and texture synthesis.

    \item We introduce correspondence-guided geometry generation through a Canonical Correspondence Encoder (CCE), which uses dense surface-coordinate supervision to provide spatial guidance for generative reconstruction.

    \item We propose geometry-guided texture generation through a geometry adapter that leverages template correspondence to incorporate reconstructed surface information into texture synthesis.
\end{itemize}

%% file: sec/2_related.tex
\begin{figure*}[t]
\centering
\includegraphics[width=\textwidth]{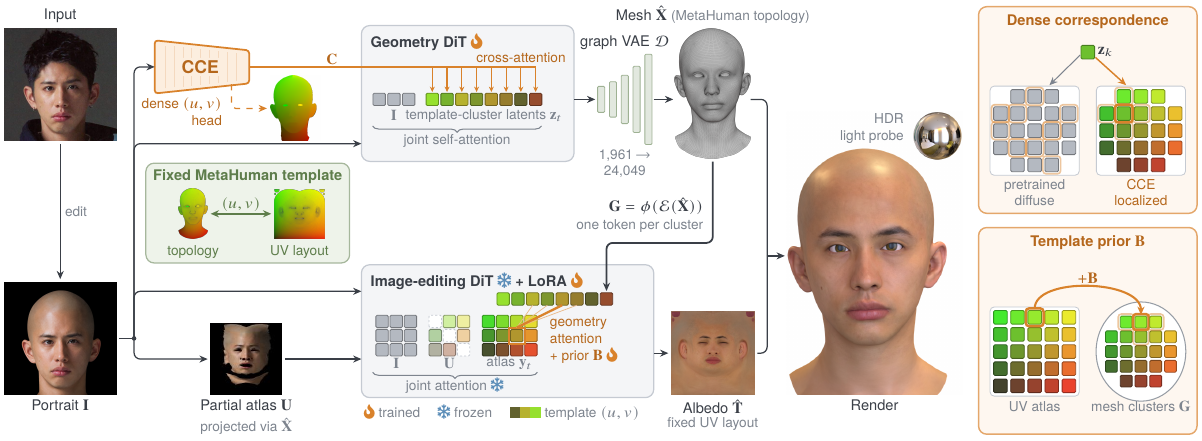}
\caption{\textbf{Overview of NERVFace.}
Both stages share a fixed MetaHuman template whose topology and UV layout correspond; red--green encodes template $(u,v)$.
Top: template-cluster latents cross-attend to features $\mathbf C$ of CCE, trained on dense $(u,v)$, and are decoded into the template topology.
Bottom: a frozen image-editing DiT with a FaceUV LoRA completes the atlas from $\mathbf I$ and $\mathbf U$, attending to the geometry memory $\mathbf G$ of the generated mesh.
Right: CCE features localize each cluster in the image; the prior $\mathbf B$ links each atlas token to the clusters at its template location.}
\label{fig:pipeline}
\end{figure*}

\section{Related Work}
\label{sec:related}

\paragraph{Native-topology face and head generation.}
Parametric face models~\cite{blanz1999morphable,flame} provide consistent topology for reconstruction methods such as DECA~\cite{deca}, MICA~\cite{mica}, and SMIRK~\cite{smirk}.
Graph and spiral mesh autoencoders~\cite{coma,neural3dmm,spiralnetpp} learn structured representations of facial geometry, while Diverse3DFace~\cite{diverse3dface} combines a graph VAE with optimization for diverse reconstructions.
General image-to-3D generators~\cite{trellis2,hunyuan3d21} produce detailed shapes without enforcing a predefined topology.
More recently, MeshLAM~\cite{meshlam} reconstructs FLAME-based geometry and texture.
Like \method, the concurrent TOPOS~\cite{topos} works in the MetaHuman topology, but conditions its flow model on generic DINOv2 features~\cite{dinov2} and generates the atlas from the portrait without the reconstructed mesh.
\method instead conditions generation on correspondence-trained features, completes the texture from portrait colors unwarped through the reconstructed mesh, and shares mesh clusters between its graph VAE and geometry adapter.

\paragraph{Dense correspondence and geometry reconstruction.}
Dense surface correspondence associates image observations with locations on a canonical template.
DensePose~\cite{densepose} and continuous surface embeddings~\cite{cse} establish such mappings for human bodies, while DenseReg~\cite{densereg} and PRNet~\cite{prnet} predict dense facial correspondences or UV position maps.
Dense landmarks~\cite{denselandmarks}, FlowFace~\cite{flowface}, Pixel3DMM~\cite{pixel3dmm}, and RealDenseFace~\cite{realdenseface} further exploit dense predictions for facial geometry estimation and model fitting.
These methods demonstrate the value of canonical surface representations for connecting image evidence to facial geometry.
Rather than using predicted coordinates as explicit fitting constraints, we train a Canonical Correspondence Encoder (CCE) with dense surface-coordinate supervision and use its intermediate representations to guide generative head reconstruction.

\paragraph{Geometry-conditioned texture synthesis.}
Facial texture reconstruction has been approached through unwarped image observations~\cite{uvgan,ostec}, learned appearance priors~\cite{ganfit,fitme}, reflectance estimation~\cite{mosar}, and UV-space generation~\cite{ffhquv,uvidm,freeuv}.
Several methods incorporate geometric information into texture synthesis.
FreeUV~\cite{freeuv} uses UV-space structural guidance, while MARCUS~\cite{marcus} combines geometric unwarping, cascaded diffusion priors, and shading constraints.
For general objects, TEXGen~\cite{texgen} combines UV-space convolutions with attention over surface point clouds, and DirectUV~\cite{directuv} uses surface-aware positional encoding to guide texture generation.
These methods illustrate different ways of using surface geometry to guide appearance synthesis.
In our framework, the predefined template establishes a known correspondence between mesh vertices and UV coordinates.
We use this correspondence to guide attention to the reconstructed surface within a frozen, UV-adapted image-editing model, complementing the appearance information provided by projected portrait colors.

%% file: sec/3_method.tex
\section{Method}
\label{sec:method}

\subsection{Framework Overview}
\label{sec:overview}
\label{sec:vae}

Given an RGB portrait $\mathbf I$, \method generates a head mesh
$\hat{\mathbf X}\in\mathbb{R}^{N\times3}$ and a full albedo
atlas $\hat{\mathbf T}$ in a predefined MetaHuman topology and
UV layout, where $N$ is the number of template vertices.
We denote the corresponding ground-truth geometry and texture
by $\mathbf X$ and $\mathbf T$.

As illustrated in \cref{fig:pipeline}, our framework consists
of two generation stages. First, a Canonical Correspondence
Encoder (CCE) learns image-to-surface associations through
dense coordinate supervision. Its representations condition
a rectified flow to generate head geometry in the predefined
topology. Second, visible portrait colors are projected
through the generated mesh to form a partial UV atlas
$\mathbf U$. A UV-adapted image-editing model generates the
full texture, guided by the portrait, partial atlas, and
geometric context from the reconstructed surface. A
geometry-attention adapter incorporates this context using
the known correspondence between the template mesh and
its UV layout.

\paragraph{Shared mesh representation.}
We construct a fixed hierarchy over the template mesh using
Graclus~\cite{graclus}, assigning each vertex $i$ to a coarse
cluster $\kappa(i)\in\{1,\dots,N_{\mathrm c}\}$. These
assignments are shared across all subjects. A hierarchical graph VAE encodes a mesh $\mathbf X$ into
cluster-level features
$\mathbf H=\mathcal E(\mathbf X)\in
\mathbb{R}^{N_{\mathrm c}\times d_e}$,
which parameterize a Gaussian posterior over mesh latents
$\mathbf z\in\mathbb{R}^{N_{\mathrm c}\times d_z}$ with mean
$\bm\mu$. The decoder $\mathcal D$ reconstructs vertex
positions from these latents (\suppref{sec:supp_vae}). The fixed mesh hierarchy provides template-cluster latents
for geometry generation. It also allows the generated mesh
to be re-encoded into cluster-level geometric features for
texture synthesis, providing a common structural reference
for both stages.

\subsection{Canonical Correspondence Encoder}
\label{sec:uvexpert}

Canonical surface coordinates associate image observations
with locations on a predefined template, providing spatial
guidance across changes in viewpoint and facial shape.
We train a Canonical Correspondence Encoder (CCE) with dense
surface-coordinate supervision to make this information
available to the geometry flow.

\paragraph{Dense correspondence supervision.}
For each valid foreground pixel $p$, we define a target
coordinate field
$\mathbf Y^{\mathrm{uv}}(p)=(u_p,v_p)$,
where $(u_p,v_p)$ identifies the corresponding location
on the template surface.
Unlike sparse landmarks, these coordinates provide dense
image-to-surface correspondence across visible facial regions. We obtain the targets by rendering training meshes from
their recorded cameras using UV-encoded materials
(\cref{fig:pipeline}).
The training data includes variations in facial shape,
appearance, viewpoint, and illumination
(\cref{sec:data_construction}).

\paragraph{Correspondence-aware representations.}
We initialize CCE from a pretrained
Sapiens2~\cite{sapiens2} backbone and attach an upsampling
head to predict the surface-coordinate field.
Training uses masked $\ell_1$ and Smooth-$\ell_1$ coordinate
losses, together with a spatial gradient loss. Our goal is not to use predicted coordinates as explicit
mesh-fitting constraints, but to learn image representations
informed by surface correspondence.
Predicted coordinates are a two-channel bottleneck: they discard
shading and silhouette cues, are discontinuous at UV seams, and
small errors become large surface offsets where the
parameterization is compressed, so fitting to them needs
test-time optimization with outlier
rejection~\cite{pixel3dmm}. We keep the correspondence-trained
features instead.
After training, we freeze CCE and extract its intermediate
patch features
$\mathbf F^{\mathrm{cce}}\in
\mathbb{R}^{N_{\mathrm p}\times d_c}$,
where $N_{\mathrm p}$ and $d_c$ are the number and dimension
of image patch features.
These representations provide correspondence-aware
conditioning for geometry generation.

\subsection{Correspondence-guided Geometry}
\label{sec:geoflow}

We generate head geometry using a conditional rectified
flow~\cite{rectifiedflow,flowmatching} in the graph VAE
latent space. The fixed mesh hierarchy organizes geometry
into template-cluster tokens, while CCE provides
correspondence-aware image conditioning.

For each training mesh, we normalize the graph VAE
posterior mean $\bm\mu$ as
$\mathbf z_0=(\bm\mu-m)/s$,
where $m$ and $s$ are fixed scalar statistics.

\paragraph{Correspondence-guided flow.}
We form the spatial condition from CCE features using
LayerNorm and learned patch-position embeddings:
\begin{equation}
\mathbf C=
\LN(\mathbf F^{\mathrm{cce}})
+\mathbf E^{\mathrm{cce}},
\label{eq:cond}
\end{equation}
where $\mathbf C\in\mathbb{R}^{N_{\mathrm p}\times d_c}$.
In addition, a frozen image VAE~\cite{qwenimage}
encodes the background-masked portrait into clean
reference tokens $\mathbf x^{\mathrm r}$. Following rectified flow, we interpolate between
the geometry latent $\mathbf z_0$ and Gaussian noise
$\bm\epsilon$:
$\mathbf z_t=(1-t)\mathbf z_0+t\bm\epsilon$.
The geometry flow predicts the velocity
$\bm\epsilon-\mathbf z_0$ by minimizing
\begin{equation}
\mathcal L_{\mathrm{RF}}=
\mathbb E_{\mathbf z_0,\bm\epsilon,t}
\left\|
v_{\theta_{\mathrm{geo}}}
(\mathbf z_t,t;\mathbf C,\mathbf x^{\mathrm r})
-(\bm\epsilon-\mathbf z_0)
\right\|_2^2.
\label{eq:rf}
\end{equation}
Here $\theta_{\mathrm{geo}}$ denotes the geometry-flow
parameters. The two image conditions provide
correspondence-aware spatial features and clean
image-latent information, respectively.

\paragraph{Conditional transformer.}
The noisy geometry latents are embedded into tokens
$\mathbf x^{\mathrm g}$ with learned mesh-cluster
position embeddings.
The transformer processes geometry and reference tokens
jointly as
$\mathbf x=[\mathbf x^{\mathrm g};\mathbf x^{\mathrm r}]$:
\begin{align}
\mathbf x
&\leftarrow \mathbf x+
\bm\gamma_1(t)\odot
\mathrm{SA}\big(\mathrm{adaLN}_1(\mathbf x,t)\big),
\\
\mathbf x^{\mathrm g}
&\leftarrow \mathbf x^{\mathrm g}+
\mathrm{CA}\big(\LN(\mathbf x^{\mathrm g}),\mathbf C\big),
\label{eq:ca}
\\
\mathbf x
&\leftarrow \mathbf x+
\bm\gamma_2(t)\odot
\mathrm{FFN}\big(\mathrm{adaLN}_2(\mathbf x,t)\big).
\end{align}
The adaptive normalization and zero-initialized gates
$\bm\gamma_1,\bm\gamma_2$ are modulated by the timestep
embedding~\cite{dit}. Joint self-attention exchanges information between
geometry and image-reference tokens, while cross-attention
allows geometry tokens to retrieve CCE features.
Importantly, the attention weights are learned rather
than determined by predicted surface coordinates.
An RMSNorm and a zero-initialized output head predict
the velocity from the updated geometry tokens. We train with logit-normal timestep sampling~\cite{sd3}.
At inference, Euler integration from $t=1$ to $t=0$
produces $\hat{\mathbf z}_0$, which is decoded as
$\hat{\mathbf X}=\mathcal D(s\hat{\mathbf z}_0+m)$.
Additional details are given in
\suppref{sec:supp_flow}.

\begin{figure*}[t]
\centering
\includegraphics[width=\textwidth]{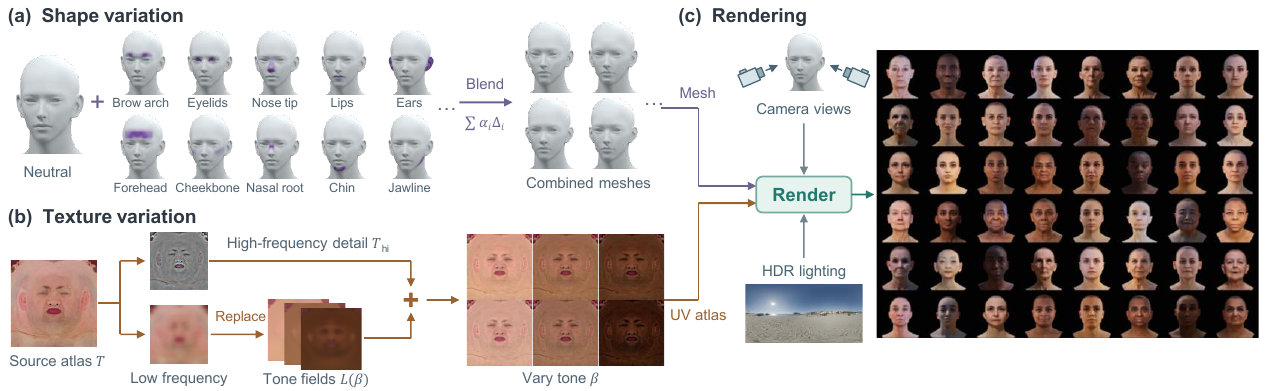}
\caption{\textbf{Dataset construction from source assets to rendered portraits.}
(a)~Violet marks the vertex displacement of ten blendshape targets; four choices of weights $\bm\alpha$ yield the meshes on the right.
(b)~Each skin-tone variant adds the original high-frequency residual (contrast enhanced) to a new smooth color field.
(c)~Viewpoint and environment lighting vary across the rendered portraits.}
\label{fig:data_construction}
\end{figure*}

\subsection{Geometry-guided Texture}
\label{sec:texture}

We generate a full albedo atlas by combining observed
appearance with geometric context from the reconstructed
mesh. The portrait and projected UV observations provide
appearance guidance, while a geometry-attention adapter
uses template correspondence to incorporate information
from the reconstructed surface.

\paragraph{Appearance conditioning.}
We first align the
reconstructed mesh $\hat{\mathbf X}$ with the input portrait
by estimating a weak-perspective camera from detected
2D facial landmarks~\cite{mediapipe} and their corresponding
mesh vertices. Using the estimated camera, we project
visible portrait colors onto the mesh surface and map
them to the predefined UV layout, producing a partial
UV atlas $\mathbf U$, which we refer to as the \emph{unwarp} (\cref{fig:pipeline}, bottom).
Local projection holes are filled as described in
\suppref{sec:supp_tex}.

We use Qwen-Image-Edit~\cite{qwenimage}, adapted for UV
texture generation through a fused FaceUV LoRA~\cite{lora}.
The portrait $\mathbf I$ and partial atlas $\mathbf U$
serve as clean references that guide full-atlas generation.
The UV-adapted appearance backbone remains frozen during
geometry-adapter training.
Using albedo atlases as targets encourages
the model to suppress illumination from the observed
appearance without a separate de-lighting stage.

\paragraph{Geometry memory.}
To provide geometric context, we re-encode the reconstructed
mesh using the frozen graph VAE, obtaining
$\hat{\mathbf H}=\mathcal E(\hat{\mathbf X})$.
For each template cluster $k$, we combine its encoded
feature $\hat{\bm h}_k$, mean vertex position
$\bar{\bm c}_k$, and normalized average surface normal
$\bar{\bm n}_k$.
A trainable MLP $\phi$ produces the geometry memory
$\mathbf G\in\mathbb R^{N_{\mathrm c}\times d_g}$:
\begin{equation}
\bm g_k=
\phi\big(
[\hat{\bm h}_k,\bar{\bm c}_k,\bar{\bm n}_k]
\big),
\qquad k=1,\dots,N_{\mathrm c}.
\label{eq:geomem}
\end{equation}
Each memory token represents a fixed template cluster
with a known correspondence to the UV atlas.

\paragraph{Correspondence-guided attention.}
We insert trainable cross-attention branches into selected
blocks of the frozen image-editing model
(\cref{fig:pipeline}).
Texture states provide queries $\mathbf Q$, while
geometry memory $\mathbf G$ supplies keys $\mathbf K_G$
and values $\mathbf V_G$. To guide texture-to-geometry attention, we construct
a bias $\mathbf B$ from the fixed template UV layout
and mesh hierarchy.
For target texture tokens $\mathcal J$ within the UV
layout, the adapter computes
\begin{align}
\mathbf A_h &=
\softmax\left(
\frac{
\mathbf Q_{\mathcal J,h}
\mathbf K_{G,h}^{\top}
}{\sqrt{d_h}}
+\mathbf B_{\mathcal J}
\right),
\notag\\
\mathbf R_{\mathcal J}
&=
\mathrm{Concat}_h
(\mathbf A_h\mathbf V_{G,h})\mathbf P_o.
\label{eq:adapter}
\end{align}
Here $h$ indexes attention heads, $d_h$ is the
per-head dimension, and $\mathbf P_o$ is the output
projection. The bias favors mesh clusters corresponding to each
atlas region and their surface neighbors: each target
token $j$ covers a nominal $16{\times}16$ texel window, whose
overlap with the mesh clusters, expanded by one edge in
the cluster adjacency graph, defines prior weights
$\widetilde W_{jk}$ and the log-bias
$B_{jk}=\log(\widetilde W_{jk}+10^{-6})$ (\cref{fig:pipeline}, right).
It is fixed across subjects and acts as a soft preference,
allowing attention to access other clusters when useful.
The resulting adapter residual is scaled by the frozen
block's timestep gate $\bm\gamma$ and added only to target
texture tokens $\bm s_j$, $j\in\mathcal J$. All adapter blocks share the same geometry memory.

We train the geometry-memory projection $\phi$ and
geometry adapters while keeping the appearance backbone
frozen. The training objective follows weighted rectified
flow:
\begin{equation}
\mathcal L_{\mathrm{tex}}=
\mathbb E_{\mathbf y_0,\bm\epsilon,t}
\left[
w(t)
\left\|
v_{\theta_{\mathrm{tex}},\psi}
(\mathbf y_t,t;\mathbf I,\mathbf U,\mathbf G)
-(\bm\epsilon-\mathbf y_0)
\right\|_2^2
\right],
\label{eq:texloss}
\end{equation}
where $\mathbf y_0$ is the VAE latent of the target
albedo atlas $\mathbf T$,
$\mathbf y_t=(1-t)\mathbf y_0+t\bm\epsilon$,
$\theta_{\mathrm{tex}}$ denotes the frozen appearance
backbone, $\psi$ the trainable
geometry-conditioning modules, and $w(t)$ the timestep weighting of \suppref{sec:supp_tex}. During training, geometry conditions are constructed
from decoded ground-truth mesh latents and recorded
cameras. At inference, they are obtained from the
generated mesh and estimated camera.
Implementation details are provided in
\suppref{sec:supp_tex}.

%% file: sec/4_experiments.tex
\section{Experiments}
\label{sec:experiments}

\begin{figure*}[t]
\centering
\includegraphics[width=\textwidth]{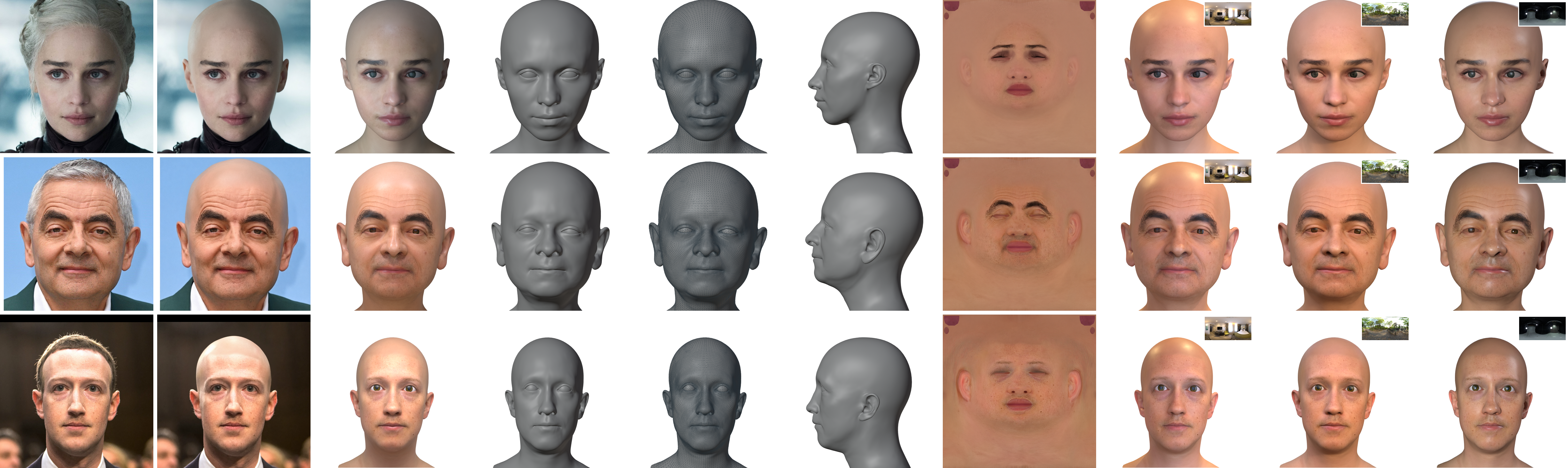}\par
\begingroup
\sffamily\fontsize{7.5}{8.2}\selectfont
\newcommand{\gallerycol}[1]{\makebox[.1\linewidth][c]{\strut#1}}%
\noindent\makebox[\linewidth][c]{\gallerycol{Input}\gallerycol{Bald portrait $\mathbf I$}\gallerycol{Input lighting}\gallerycol{Mesh}\gallerycol{Wireframe}\gallerycol{Side view}\gallerycol{Albedo $\hat{\mathbf T}$}\gallerycol{Interior HDR}\gallerycol{Forest HDR}\gallerycol{Studio HDR}}\par
\endgroup
\caption{\textbf{Results on in-the-wild portraits.} Per row: photograph, bald portrait $\mathbf I$, render under lighting fitted to the photograph, mesh (frontal, wireframe, and side views), albedo atlas, and its renders under three HDR maps (insets).}
\label{fig:outputs}
\end{figure*}

\subsection{Training Dataset Construction}
\label{sec:data_construction}

We construct training data by varying facial geometry and appearance within a shared MetaHuman topology and UV layout. This preserves known correspondence between rendered portraits, head surfaces, and texture atlases, providing supervision for both generation stages and CCE (\cref{fig:data_construction}).

\paragraph{Geometry augmentation.}
We augment each base head $\mathbf X_0$ using blendshape offsets grouped into nine facial regions, including the brows, eyes, nose, mouth, and chin. For each region $r$, we select three offsets $\bm\Delta_{rj}$ and sample their coefficients within artist-defined bounds. If the sampled weights in a region sum to more than one, they are normalized by their sum. The augmented geometry is given by
\begin{equation}
\mathbf X(\bm\alpha)=
\mathbf X_0+
\sum_r\sum_{j=1}^{3}
\alpha_{rj}\bm\Delta_{rj}.
\label{eq:shape_augmentation}
\end{equation}
This produces diverse facial shapes while preserving mesh connectivity, vertex indexing, and UV correspondence (\cref{fig:data_construction}a). We generate 193{,}887 mesh instances from 387 base characters. CCE, the geometry flow and the texture adapter use 347 training characters; the remaining 40 are held out. The graph VAE uses a separate mesh-instance split across all 387 characters and receives no portraits. The geometry flow additionally uses LYHM~\cite{lyhm,lyhm_ijcv2020} and FaceScape~\cite{facescape} scans registered to the template (158 portrait--scan pairs), with no subject overlap between training and real-scan evaluation. All other components are trained or adapted using only our synthetic dataset (\suppref{sec:supp_data}).

\paragraph{Texture augmentation.}
To vary skin appearance while retaining fine details, we decompose a source albedo atlas $\mathbf T$ into a low-frequency component and a high-frequency residual using Gaussian filtering $\mathcal S_\sigma$. A two-parameter skin-tone model produces a smooth color field $\mathbf L(\bm\beta)$, which replaces the low-frequency component:
\begin{equation}
\begin{aligned}
\mathbf T_{\mathrm{hi}}
&=\mathbf T-\mathcal S_\sigma(\mathbf T),\\
\mathbf T_{\bm\beta}
&=\operatorname{clip}_{[0,1]}
\big(\mathbf L(\bm\beta)+\mathbf T_{\mathrm{hi}}\big).
\end{aligned}
\label{eq:skintone_augmentation}
\end{equation}
This changes overall skin coloration while preserving local texture details (\cref{fig:data_construction}b). Additional atlas edits vary eyebrows and makeup. In total, we generate 127{,}316 textures from 420 artist-created and scan-derived atlases.

\paragraph{Rendering.}
We render selected geometry--texture pairs in Blender under varied viewpoints and HDR environment lighting (\cref{fig:data_construction}c). Alongside each RGB portrait, we render a pixel-aligned UV coordinate field as dense supervision for CCE (\cref{fig:pipeline}, top). The corresponding mesh and complete albedo atlas provide training targets for geometry and texture generation, respectively.

\subsection{Evaluation Protocol}
\label{sec:setup}
We evaluate \method on synthetic and real-scan test sets. All compared methods receive the same masked, hair-removed portraits as input.

The geometry evaluation sets (\cref{tab:geo}) contain 200 synthetic subjects and 300 real-scan subjects. For the real-scan sets, we randomly sample 200 and 100 subjects from LYHM~\cite{lyhm,lyhm_ijcv2020} and FaceScape~\cite{facescape}, respectively, each with a textured 3D scan from which the input portrait is rendered. The synthetic subjects are frontal renders under randomly sampled HDR environment lighting, derived from the 40 held-out characters described in \cref{sec:data_construction}. When computing metrics, the reconstructed meshes are similarity-aligned to the ground truth using seven facial landmarks followed by ICP, without non-rigid deformation. We report S2M, the mean distance from points sampled on the ground-truth facial region to the predicted surface, in percent of the outer-eye-corner distance (OCD); NC, the mean absolute cosine between ground-truth and predicted normals at closest-point pairs; and R@1\%, the percentage of samples within 1\%\,OCD. For topology-specific ablations, we additionally report vertex-to-vertex error (V2V), averaged over all 24{,}049 corresponding vertices, and bidirectional point-to-triangle Chamfer distance (CD). Both are computed in canonical coordinates without ICP and reported in units of $10^{-3}$.

The texture evaluation uses the same synthetic and real subjects as the geometry evaluation. For 200 synthetic subjects, we render input portraits under randomly sampled HDR environment lighting and compare the results with unlit renders of the ground-truth albedo on the ground-truth mesh. For the 300 real subjects, the scanned textures, which contain baked-in environmental lighting, are used for both input rendering and metric computation; each method's output texture is rendered without shading on its reconstructed mesh, so any lighting baked into the predicted texture is kept, whereas FFHQ-UV outputs only a lighting-normalized texture. All texture metrics are computed on rendered images of the textured heads rather than directly in UV space. We measure PSNR, SSIM~\cite{ssim}, and LPIPS~\cite{lpips} over the ground-truth face region. These metrics evaluate the complete textured reconstruction and therefore also reflect geometry and alignment errors.

\begin{figure*}[t]
\centering
\includegraphics[width=\textwidth,trim=0 0 0 16bp,clip]{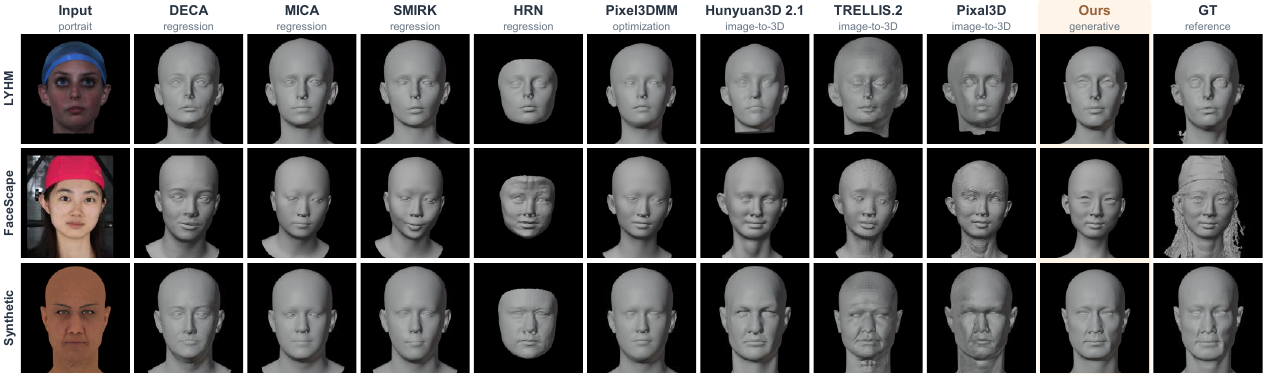}\par
\begingroup
\sffamily\fontsize{6}{7}\selectfont
\newcommand{\geoqualcol}[1]{\makebox[.0894086\linewidth][c]{\strut#1}}%
\noindent\hbox to\linewidth{\hspace*{.0147984\linewidth}\geoqualcol{Input}\geoqualcol{DECA}\geoqualcol{MICA}\geoqualcol{SMIRK}\geoqualcol{HRN}\geoqualcol{Pixel3DMM}\geoqualcol{Hunyuan3D 2.1}\geoqualcol{TRELLIS.2}\geoqualcol{Pixal3D}\geoqualcol{\textbf{Ours}}\geoqualcol{GT}\hss}\par
\endgroup
\caption{\textbf{Qualitative geometry comparison.}
Rows show examples from LYHM~\cite{lyhm}, FaceScape~\cite{facescape}, and the synthetic dataset. The synthetic example is a mesh excluded from flow training whose base character was used for training; \cref{tab:geo} reports results on held-out characters.}
\label{fig:geo_qual}
\end{figure*}

\begin{table}[t]
\centering
\caption{\textbf{Geometry comparison.}
\method achieves high geometric accuracy while preserving standardized MetaHuman topology across all reconstructed heads. Bold: best; underline: second.}
\label{tab:geo}
\scriptsize
\setlength{\tabcolsep}{1.2pt}
\resizebox{\linewidth}{!}{%
\begin{tabular}{@{}l ccc ccc ccc@{}}
\toprule
& \multicolumn{3}{c}{LYHM~\cite{lyhm}}
& \multicolumn{3}{c}{FaceScape~\cite{facescape}}
& \multicolumn{3}{c}{Synthetic} \\
\cmidrule(lr){2-4}\cmidrule(lr){5-7}\cmidrule(l){8-10}
Method
& S2M$\downarrow$ & NC$\uparrow$ & R@1\%$\uparrow$
& S2M$\downarrow$ & NC$\uparrow$ & R@1\%$\uparrow$
& S2M$\downarrow$ & NC$\uparrow$ & R@1\%$\uparrow$ \\
\midrule
DECA~\cite{deca}
& 1.282 & 0.950 & 48.8
& 1.660 & 0.930 & 38.7
& 1.896 & 0.932 & 35.5 \\
MICA~\cite{mica}
& \second{0.986} & \second{0.956} & 60.4
& 0.985 & 0.951 & 60.6
& 1.763 & \second{0.934} & 37.3 \\
SMIRK~\cite{smirk}
& 1.280 & 0.947 & 49.2
& 1.404 & 0.930 & 45.0
& 1.903 & 0.932 & 35.1 \\
HRN~\cite{hrn}
& 1.221 & 0.949 & 50.8
& 1.322 & 0.948 & 46.3
& 2.446 & 0.931 & 27.4 \\
Pixel3DMM~\cite{pixel3dmm}
& 1.224 & 0.951 & 50.8
& \best{0.867} & \second{0.954} & \best{66.1}
& 1.698 & 0.933 & 38.9 \\
Hunyuan3D 2.1~\cite{hunyuan3d21}
& 1.097 & 0.954 & \second{60.6}
& 1.347 & 0.936 & 54.0
& \second{1.275} & 0.930 & \best{57.1} \\
TRELLIS.2~\cite{trellis2}
& 1.267 & 0.942 & 49.9
& 1.310 & 0.939 & 48.9
& 1.970 & 0.914 & 34.3 \\
Pixal3D~\cite{pixal3d}
& 1.371 & 0.921 & 46.5
& 1.211 & 0.927 & 50.6
& 1.899 & 0.911 & 35.7 \\
\midrule
\method (ours)
& \best{0.978} & \best{0.962} & \best{61.3}
& \second{0.892} & \best{0.964} & \second{64.3}
& \best{1.197} & \best{0.960} & \second{53.0} \\
\bottomrule
\end{tabular}}
\end{table}

\begin{table}
\centering
\caption{\textbf{Texture comparison.}
\method outperforms all baselines across all three metrics on both real and synthetic subjects. $^\ast$Per-image optimization. Bold: best; underline: second.}
\label{tab:tex}
\footnotesize
\setlength{\tabcolsep}{2pt}
\resizebox{\linewidth}{!}{%
\begin{tabular}{@{}l ccc ccc@{}}
\toprule
& \multicolumn{3}{c}{Real} & \multicolumn{3}{c@{}}{Synthetic} \\
\cmidrule(lr){2-4}\cmidrule(l){5-7}
Method & PSNR$\uparrow$ & SSIM$\uparrow$ & LPIPS$\downarrow$ & PSNR$\uparrow$ & SSIM$\uparrow$ & LPIPS$\downarrow$ \\
\midrule
HRN~\cite{hrn} & 23.63 & 0.783 & 0.281 & 17.74 & 0.783 & 0.350 \\
UV-IDM~\cite{uvidm} & 23.79 & \second{0.796} & 0.319 & 18.29 & 0.803 & 0.374 \\
FreeUV~\cite{freeuv} & 22.59 & 0.781 & 0.316 & 18.94 & 0.811 & 0.376 \\
FFHQ-UV$^\ast$~\cite{ffhquv} & 21.25 & 0.774 & 0.273 & 21.73 & \second{0.833} & \second{0.296} \\
MARCUS~\cite{marcus} & \second{24.52} & 0.783 & \second{0.205} & \second{22.28} & 0.762 & 0.314 \\
\midrule
\method (ours) & \best{26.49} & \best{0.833} & \best{0.149} & \best{24.67} & \best{0.876} & \best{0.125} \\
\quad w/o unwarp & 23.91 & 0.779 & 0.233 & 23.48 & 0.865 & 0.180 \\
\quad w/o adapter & 24.93 & 0.806 & 0.183 & 24.13 & 0.869 & 0.153 \\
\bottomrule
\end{tabular}}
\end{table}

\subsection{Results}
\label{sec:exp_geo}

\noindent \textbf{Geometry.}
\Cref{fig:outputs} shows complete outputs for in-the-wild portraits. \Cref{tab:geo} compares parametric regressors, dense fitting methods, and image-to-3D generators, with qualitative results in \cref{fig:geo_qual}.
\method ranks first or second across all metrics on the three datasets.
On LYHM, it achieves the lowest S2M (0.978) and highest R@1\% (61.3), with small margins over MICA~\cite{mica} and Hunyuan3D~2.1~\cite{hunyuan3d21}.
On FaceScape, Pixel3DMM~\cite{pixel3dmm} performs slightly better in S2M and R@1\%.
On the synthetic set, Hunyuan3D~2.1 achieves higher R@1\% but has a larger mean surface error.
Compared with general-purpose 3D generators such as Hunyuan3D~2.1 and TRELLIS.2, \method achieves lower surface errors and higher normal consistency across all three datasets.
Moreover, our meshes retain the standard MetaHuman topology and its edge loops, avoiding the need for additional retopology.
MICA and Pixel3DMM were trained on LYHM and FaceScape scans, respectively, which may include our test subjects.
In contrast, the real scans used to train \method are subject-disjoint from the evaluation sets.
Finally, rigidly aligned surface distances can favor overly smooth geometry.
We therefore also report normal consistency, where \method performs best on all three datasets.

\noindent \textbf{Texture.}
\Cref{tab:tex} compares texture reconstruction on real and synthetic subjects. Unlike methods that rely on separate illumination normalization or decomposition stages, such as MARCUS and FFHQ-UV, \method directly generates complete albedo atlases in the predefined UV layout, suitable for downstream rendering and editing.

For real-scan evaluation, the reference textures contain baked-in lighting. To enable a more direct appearance comparison with methods producing shaded textures, we additionally train a separate LoRA to generate textures with baked-in lighting and use these outputs for the real-scan evaluation. This auxiliary LoRA does not affect our main albedo generation pipeline. Under this setting, \method achieves the best results across all three metrics (PSNR 26.49, SSIM 0.833, and LPIPS 0.149), outperforming the strongest baseline, MARCUS, by 1.97 dB in PSNR.

On the synthetic set, the reference images are rendered from the ground-truth albedo without shading, requiring methods to recover textures independent of the input illumination.
MARCUS, FFHQ-UV, and \method are evaluated using their predicted albedo, while the remaining methods retain input lighting in their textures.
\method achieves the best results across all three metrics (PSNR 24.67, SSIM 0.876, and LPIPS 0.125), outperforming the strongest baseline by 2.39 dB in PSNR.
Removing either the unwarping or adapter module consistently degrades performance on both datasets.

\begin{figure}[t]
\centering
\includegraphics[width=\linewidth]{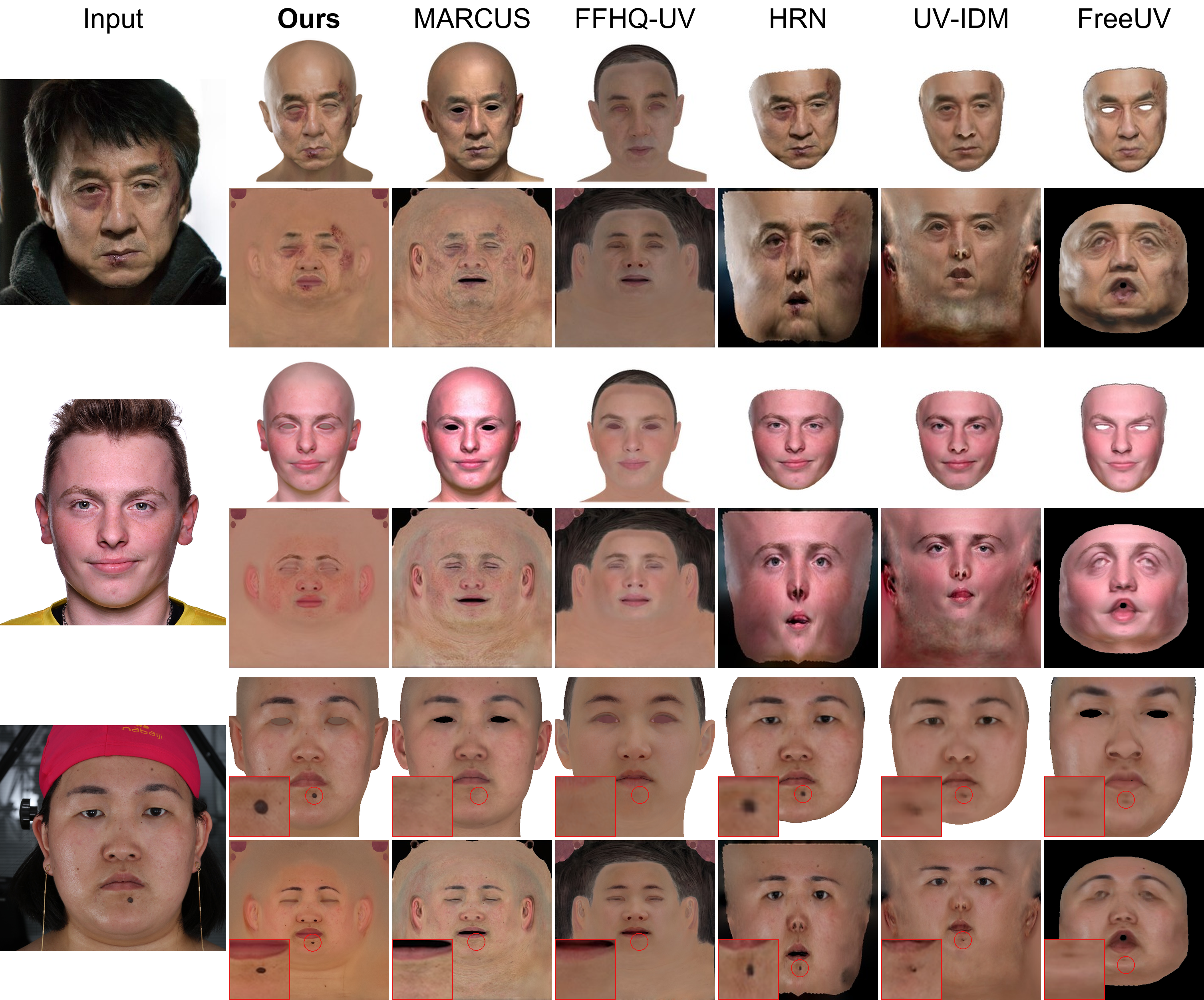}
\caption{\textbf{Qualitative appearance comparison} on in-the-wild portraits and one FaceScape subject.
Top: renderings from the input viewpoint.
Bottom: completed UV atlases (albedo for ours, MARCUS, and FFHQ-UV; textures with baked-in lighting for the remaining methods).}
\label{fig:tex_cmp}
\end{figure}

\begin{table}[!tb]
\centering
\caption{\textbf{Geometry ablations.}
Whole-head errors ($\times10^{-3}$, no ICP) on 2{,}000 cases from 40 held-out meshes. Neither CCE nor the geometry flow was trained on these base characters. All variants share the flow-training budget (\suppref{sec:supp_flow}); bold marks the lowest mean.}
\label{tab:abl_geo}
\setlength{\tabcolsep}{6pt}
\resizebox{\linewidth}{!}{%
\begin{tabular}{l cc}
\toprule
Variant & V2V$\downarrow$ & CD$\downarrow$ \\
\midrule
DINOv3 only & 18.89 & 9.96 \\
Sapiens2 Pretrain only & 18.13 & 9.81 \\
Sapiens2 Normal only & 18.07 & 9.63 \\
Sapiens2 Pointmap only & 17.93 & 9.66 \\
\midrule
CCE only (ours) & \best{16.88} & \best{9.37} \\
Sapiens2 Pretrain + CCE & 17.26 & 9.50 \\
Sapiens2 Pretrain + Normal + CCE & 17.33 & 9.46 \\
\bottomrule
\end{tabular}}
\end{table}

\subsection{Ablations}
\label{sec:ablation}

\noindent \textbf{Geometry.}
We compare feature sources at a fixed flow-training budget, keeping the mesh representation and RGB latent input unchanged (\cref{tab:abl_geo}; \suppref{sec:supp_flow}).
On held-out characters, CCE reduces mean V2V from 18.13 to 16.88 (6.9\%) and mean CD from 9.81 to 9.37 (4.5\%) relative to its pretrained Sapiens2 backbone.
Normal and pointmap features yield smaller reductions, while DINOv3~\cite{dinov3} gives higher errors than pretrained Sapiens2.
Both multi-stream variants improve on pretrained Sapiens2, but neither matches CCE alone on either metric.

\noindent \textbf{Texture.}
We ablate the unwarp and geometry adapter while keeping the reconstructed mesh, estimated camera, and sampling noise fixed (\cref{tab:tex}).
Without the unwarp, the model relies solely on the input portrait for appearance information.
Without the adapter, the model still receives the unwarped texture $\mathbf U$ and portrait, but loses geometry-aware conditioning.

Removing either component degrades performance across all metrics.
Without the unwarp, PSNR drops by 2.58 and 1.19\,dB on the real and synthetic sets, respectively, while LPIPS increases by 0.084 and 0.055.
This shows the importance of directly projected colors for preserving appearance details.
Removing the adapter leads to smaller PSNR drops of 1.56 and 0.54\,dB.
As shown in \cref{fig:abl_tex_qual}, the adapter mainly sharpens the eyebrow and lip boundaries, which has a limited effect on global metrics.
Since this ablation removes both the geometry memory and template bias $B_{jk}$, the improvement reflects their combined contribution.

%% file: sec/5_conclusion.tex
\begin{figure}[t]
\centering
\includegraphics[width=\columnwidth]{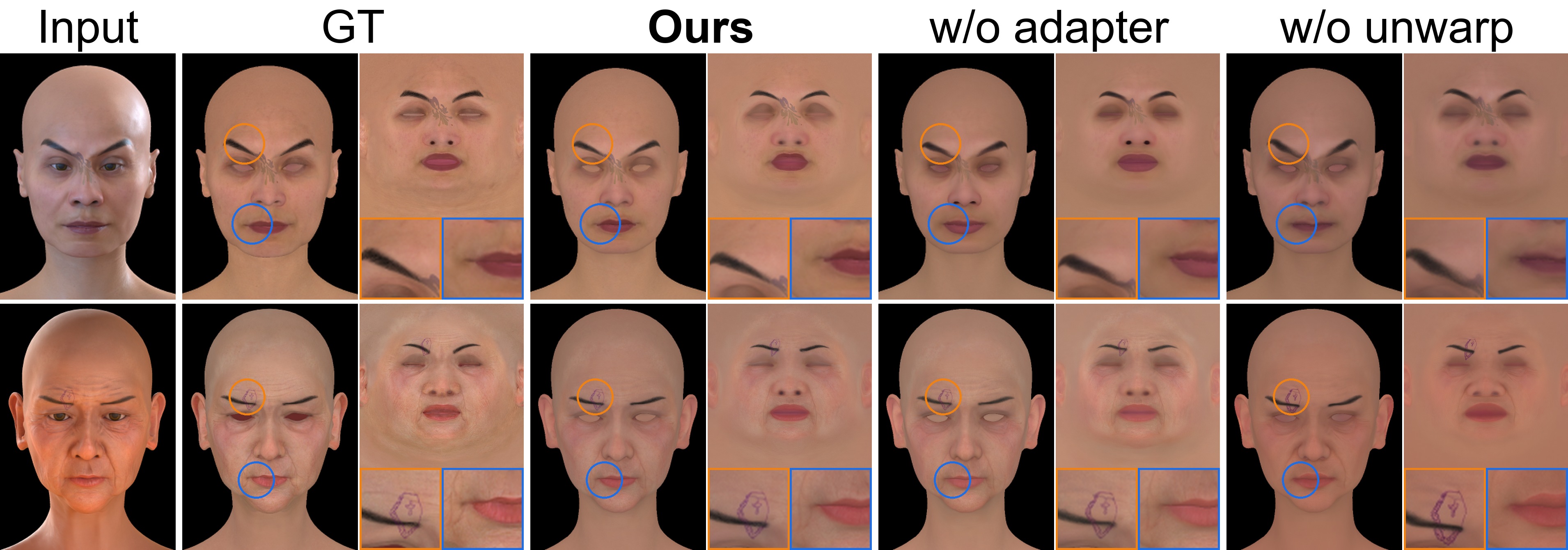}
\caption{\textbf{Qualitative texture ablation} on synthetic test subjects: render, face region of the UV atlas, and enlargements of the eyebrow (orange) and lip edge (blue). Without the unwarp, these boundaries blur and shift; without the adapter, they are less sharp.}
\label{fig:abl_tex_qual}
\end{figure}

\section{Conclusion}
\label{sec:conclusion}

We presented \method, a framework for reconstructing textured 3D heads in a predefined topology from a single image.
Unlike methods that focus primarily on geometric accuracy or visual appearance, \method produces detailed head meshes with standardized MetaHuman topology and complete, relightable UV textures, so that the reconstructed heads share the template's topology and UV interface, a prerequisite for downstream editing and rigging.
Our approach builds on dense surface correspondence to improve geometry generation and to guide texture synthesis with the reconstructed mesh.
Experiments demonstrate competitive geometric accuracy, improved texture fidelity, and the effectiveness of our correspondence-aware design.
Together, these results show that single-image head reconstruction can combine image fidelity with the fixed topology and UV layout of a template asset.

%% file: sec/6_llm_statement.tex
\section*{LLM Usage Statement}
We used ChatGPT to improve English writing and refine the clarity and organization of technical descriptions. Claude was used to assist with figure layouts and visual presentation. All AI-assisted content was carefully reviewed and revised by the authors, who take full responsibility for the final paper and supplementary material.

%% file: sec/X_suppl.tex
\renewcommand{\dblfloatpagefraction}{0.6}
\renewcommand{\floatpagefraction}{0.6}
\newcommand{\suppContentsEntry}[3]{%
  \hyperref[#1]{\textbf{#2}} & #3 & \hyperref[#1]{\pageref*{#1}}\\}
\twocolumn[{%
\centering
{\Large\textbf{Appendix}\par}
\vspace{1.0em}
\begin{minipage}{\textwidth}
{\raggedright
\noindent\textbf{Contents}\par\smallskip
\small
\noindent Sections~\ref*{sec:supp_vae}--\ref*{sec:supp_tex} describe the model implementation; Sections~\ref*{sec:supp_data} and~\ref*{sec:supp_results} present the evaluation protocols and additional results.\par\smallskip
\setlength{\tabcolsep}{0pt}%
\begin{tabular*}{\linewidth}{@{\extracolsep{\fill}}llr@{}}
\suppContentsEntry{sec:supp_vae}{A. Shared surface representation}{Mesh hierarchy, graph VAE architecture, training, reconstruction accuracy.}
\suppContentsEntry{sec:supp_flow}{B. Geometry generation}{CCE training, geometry flow, feature sources for Table~\ref*{tab:abl_geo}.}
\suppContentsEntry{sec:supp_mesh_to_uv}{C. From mesh to UV}{Correspondence construction and geometry adapter.}
\suppContentsEntry{sec:supp_tex}{D. Texture synthesis}{Base model, camera and unwarping, adapter, flow schedule.}
\suppContentsEntry{sec:supp_data}{E. Data and evaluation protocols}{Data, splits, geometry, alignment, texture, TOPOS comparison.}
\suppContentsEntry{sec:supp_results}{F. Results and applications}{Full metrics, blendshapes and rig, real-photo preprocessing and fidelity.}
\suppContentsEntry{sec:supp_impact}{Broader impact}{Consent, misuse, provenance of likeness reconstruction.}
\suppContentsEntry{fig:wild}{Portrait gallery}{Reconstructions of 42 real portraits.}
\end{tabular*}\par}
\end{minipage}
\par\medskip
}]

\setcounter{table}{2}
\section{Shared surface representation}
\label{sec:supp_vae}
Section~\ref{sec:vae} introduces a graph VAE as the shared representation for head geometry. This section describes how the mesh hierarchy is built and how the VAE is trained. Table~\ref{tab:supp_vae} reports its reconstruction error on validation meshes.

\begin{figure*}[!t]
\centering
\includegraphics[width=\textwidth]{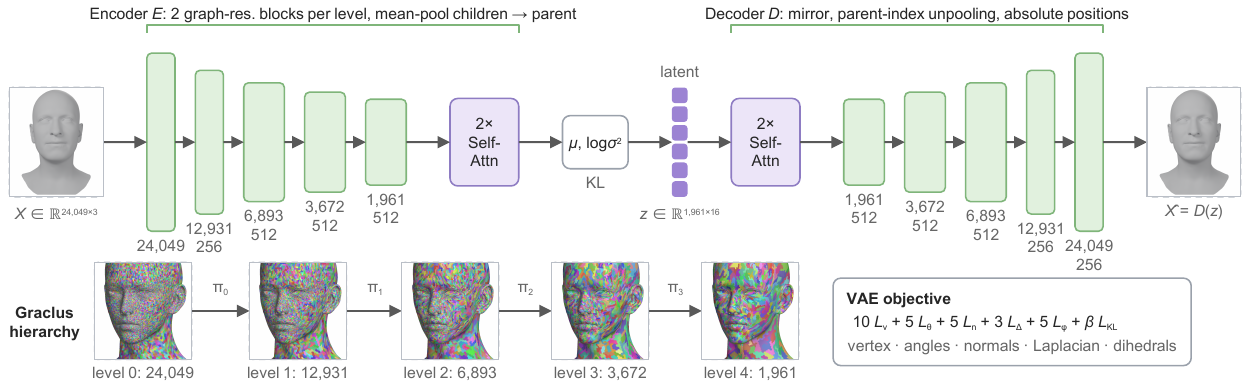}
\caption{\textbf{Hierarchical graph VAE.} Top: the encoder pools features along the mesh hierarchy; the decoder unpools through the same parent maps. Bar height scales with the square root of the node count and width with the channel count. Bottom: template vertices colored by cluster at each level.}
\label{fig:supp_vae}
\end{figure*}

\paragraph{Hierarchy.}
The finest level has 24{,}049 vertices and 48{,}004 triangles (\cref{fig:supp_vae,tab:supp_hier}).
The Graclus coarsening (\cref{sec:vae}) visits vertices in a fixed random order (seed 0 for level 0, incremented per level) and merges each unmatched vertex with its unmatched neighbor of largest weight $1/\deg(u)+1/\deg(v)$; vertices without unmatched neighbors form singleton clusters.
Edges of the next level are obtained by contracting the current edges through the parent map and removing self-loops and duplicates.
Unlike the point-cloud encoder and voxel-grid decoder in TOPOS~\cite{topos}, both halves of our VAE operate on this graph.

\begin{table}[h]
\centering
\caption{Mesh hierarchy and VAE widths.}
\label{tab:supp_hier}
\resizebox{\linewidth}{!}{%
\begin{tabular}{l ccccc}
\toprule
Level & 0 & 1 & 2 & 3 & 4 \\
\midrule
\#Nodes & 24{,}049 & 12{,}931 & 6{,}893 & 3{,}672 & 1{,}961 \\
\#Edges & 72{,}052 & 38{,}721 & 20{,}619 & 10{,}967 & 5{,}850 \\
Width & 256 & 256 & 512 & 512 & 512 \\
Res.\ blocks & 2 & 2 & 2 & 2 & 2 \\
\bottomrule
\end{tabular}}
\end{table}

\paragraph{Architecture.}
In the notation of the main paper, the encoder feature width is $d_e=512$ and the latent width is $d_z=16$. The VAE has no learned positional embedding.
Each residual block applies LayerNorm, GraphConv, GELU, LayerNorm, and GraphConv with mean aggregation, and adds a linear projection of its input.
The bottleneck uses two pre-norm self-attention blocks (8 heads, width 512, MLP ratio 4) in both encoder and decoder.
The output head is initialized with zero bias and weights drawn from a Gaussian with standard deviation $10^{-3}$.

\paragraph{Training.}
The VAE is trained on raw vertex coordinates with
$\mathcal{L}=10\,\mathcal{L}_{\mathrm{v}}+5\,\mathcal{L}_{\mathrm{ang}}+5\,\mathcal{L}_{\mathrm{n}}+3\,\mathcal{L}_{\Delta}+5\,\mathcal{L}_{\mathrm{dih}}+\beta\,\mathcal{L}_{\mathrm{KL}}$,
where $\mathcal{L}_{\mathrm{v}}$ is the vertex MSE, $\mathcal{L}_{\mathrm{ang}}$ the $\ell_1$ difference of the cosines of all triangle angles, $\mathcal{L}_{\mathrm{n}}$ the mean $1-\cos$ between predicted and ground-truth face normals, $\mathcal{L}_{\Delta}$ the mean Euclidean norm of the difference of uniform (umbrella) Laplacians, and $\mathcal{L}_{\mathrm{dih}}$ the $\ell_1$ difference of signed per-edge dihedral angles normalized by $\pi$; $\beta$ is warmed up linearly from $0$ to $10^{-4}$ over the first 2{,}000 steps.
We use AdamW (lr $6{\times}10^{-4}$, weight decay $0.01$) with 1{,}000 warm-up steps and cosine decay to $10^{-5}$, gradient clipping at 0.5, and a batch of 32 meshes on each of 32 accelerators (global batch 1{,}024) for 100k steps.
With probability 0.5, we replace a training mesh with a uniformly Laplacian-smoothed copy (1--4 iterations, step $\lambda\sim\mathcal{U}[0.1,0.5]$).
Latent statistics used for normalization are $m={-0.0392}$ and $s=2.0025$.

\paragraph{Reconstruction accuracy.}
\Cref{tab:supp_vae} evaluates the decoded posterior mean. For scale, a head spans about $1.4{\times}1.8{\times}1.1$ canonical units.
\begin{table}[h]
\centering
\caption{VAE reconstruction on a fixed subset of 200 validation meshes (canonical units, no alignment). V2V is reported $\times10^{-3}$; NC is the mean cosine between corresponding face normals.}
\label{tab:supp_vae}
\setlength{\tabcolsep}{4pt}
\begin{tabular}{l cc}
\toprule
& V2V$\downarrow$ & NC$\uparrow$ \\
\midrule
Ours (posterior mean) & 1.90 & 0.9985 \\
\bottomrule
\end{tabular}
\end{table}

\section{Geometry generation}
\label{sec:supp_flow}
Sections~\ref{sec:uvexpert} and~\ref{sec:geoflow} introduce CCE and the geometry flow conditioned on its features. This section describes CCE training, the flow architecture and the geometry ablation. Table~\ref{tab:supp_feature_sources} identifies the feature layers used in each backbone.

\par\smallskip\noindent\begin{minipage}{\columnwidth}
\centering
\edef\suppPreviousTable{\number\value{table}}
\setcounter{table}{1}
\captionsetup{type=table}
\caption{\textbf{Feature sources for the geometry ablation in Table~\ref{tab:abl_geo}.} Multi-source variants concatenate tokens from the selected backbones.}
\label{tab:supp_feature_sources}
\setcounter{table}{\suppPreviousTable}
\small
\setlength{\tabcolsep}{3pt}
\begin{tabular}{@{}ll@{}}
\toprule
Backbone & Extracted features \\
\midrule
DINOv3 H+/16~\cite{dinov3} & Generic image features \\
Pretrained Sapiens2 & Block 36 \\
Sapiens2 normal estimator & Block 28 \\
Sapiens2 pointmap estimator & Block 22 \\
CCE (correspondence-trained Sapiens2) & Block 32 \\
\bottomrule
\end{tabular}
\end{minipage}\par\smallskip

\paragraph{Canonical Correspondence Encoder.}
\label{sec:supp_uvexpert}
For dense UV supervision~\cite{flowface,pixel3dmm,decomr,lhm}, CCE takes $512^2$ portraits. Its Sapiens2-1B backbone (40 blocks, width 1536, patch size 16) feeds four upsampling stages (768/512/256/128 channels), three $3{\times}3$ convolutions (64/32/16 channels), and a sigmoid. We exclude zero-UV background pixels from all losses.
The loss is a masked $\ell_1$ (weight 1) plus a Smooth-$\ell_1$ (weight 1, $\beta=0.05$) on the UV values and an $\ell_1$ loss on their spatial gradients (weight 2).
Training uses photometric distortion, random down--up sampling, rescaling, cropping, Gaussian blur, and noise, but no horizontal flips, which would mirror the UV layout.
We train for 24k iterations with AdamW (weight decay 0.1, layer-wise decay 0.9), gradient clipping at 4, bf16 precision, and 16 images per device.

\paragraph{Feature extraction and flow architecture.}
We cache the frozen CCE features before flow training (\cref{tab:supp_hparams}) and decode sampled mesh latents in FP32.

\begin{table}[h]
\centering
\caption{Geometry flow hyperparameters.}
\label{tab:supp_hparams}
\small
\resizebox{\linewidth}{!}{%
\begin{tabular}{l l}
\toprule
Blocks / width / heads / FFN & 32 / 1024 / 16 / 4$\times$ \\
Geometry tokens & 1{,}961 $\times$ 16 \\
RGB tokens & 1{,}024 ($16{\times}64{\times}64$, $2{\times}2$ patches) \\
Feature tokens & 1{,}024, width 1536 \\
Feature layers & CCE block 32 \\
Positional encoding & learned absolute (no RoPE) \\
Timesteps / sampling & 3{,}000 / logit-normal \\
Optimizer & AdamW, lr $10^{-4}$, wd 0.01 \\
Schedule & 500 warm-up steps, cosine decay \\
Batch / grad.\ clip / EMA & 32 / 0.3 / 0.9999 \\
Precision & bf16 \\
Inference & 50 Euler steps \\
\bottomrule
\end{tabular}}
\end{table}

\paragraph{Geometry ablation protocol.}
\label{sec:supp_geo_ablation}
We train all flow variants in \cref{tab:abl_geo} on the same synthetic portrait--mesh pairs, with a fixed graph VAE, template, RGB latent input, training seed and learning-rate schedule. Each variant receives 12{,}336 optimizer updates at an effective batch size of 32. We evaluate its EMA checkpoint on the same 2{,}000 held-out cases, using batch size 16 and seed 42.
For the whole-head metrics defined in \cref{sec:setup}, CD averages the two directional point-to-triangle distances without squaring them.
The feature backbones and extraction layers are listed in \cref{tab:supp_feature_sources}.
The normal and pointmap controls are single-source models that replace CCE with the corresponding backbone; multi-source variants concatenate the source tokens, each with its own LayerNorm and embedding (\cref{eq:cond}).
Single-source variants compare feature backbones at the same flow-training budget; concatenating sources also changes token count and capacity.

\begin{figure*}[!t]
\centering
\includegraphics[width=0.90\textwidth]{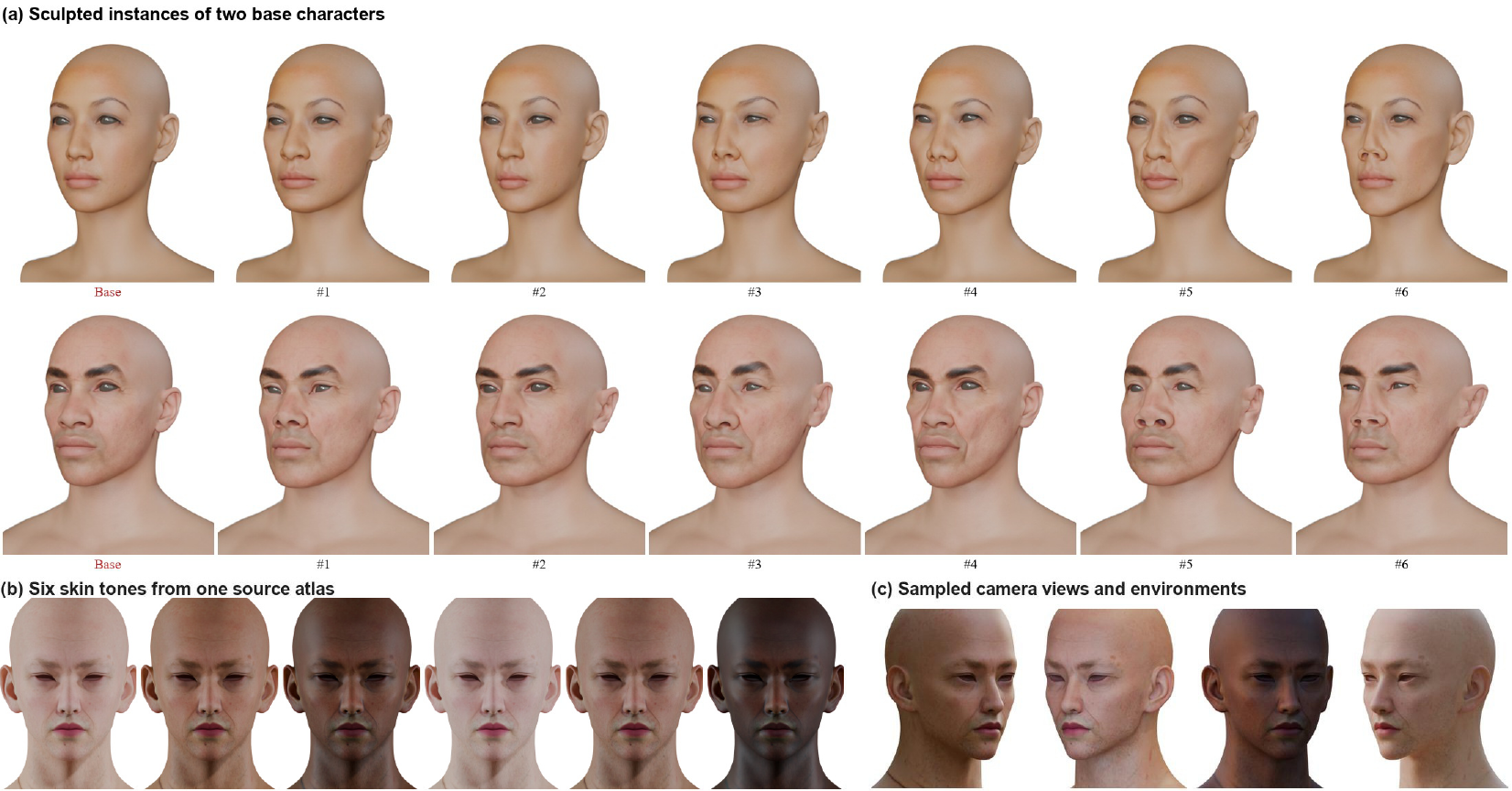}
\caption{\textbf{Synthetic data variation.} (a) Sculpted mesh instances of two base characters (base at left), each rendered with one albedo atlas under the same camera and lighting; all instances share the template topology and UV layout. (b) One shape rendered with six skin-tone fields applied to the same source atlas. (c) Sampled camera views and HDR environments.}
\label{fig:supp_datavar}
\end{figure*}

\Cref{fig:supp_datavar} illustrates the variation in shape, appearance, and viewing conditions in the synthetic data used for this ablation.

\section{From mesh to UV}
\label{sec:supp_mesh_to_uv}
Section~\ref{sec:texture} introduces the template bias used in geometry-guided attention (\cref{eq:adapter}). This section derives the bias from the overlap between UV tokens and mesh clusters, quantifies coverage and shows how each attention head uses the bias.

\begin{figure*}[p]
\centering
\includegraphics[width=\textwidth]{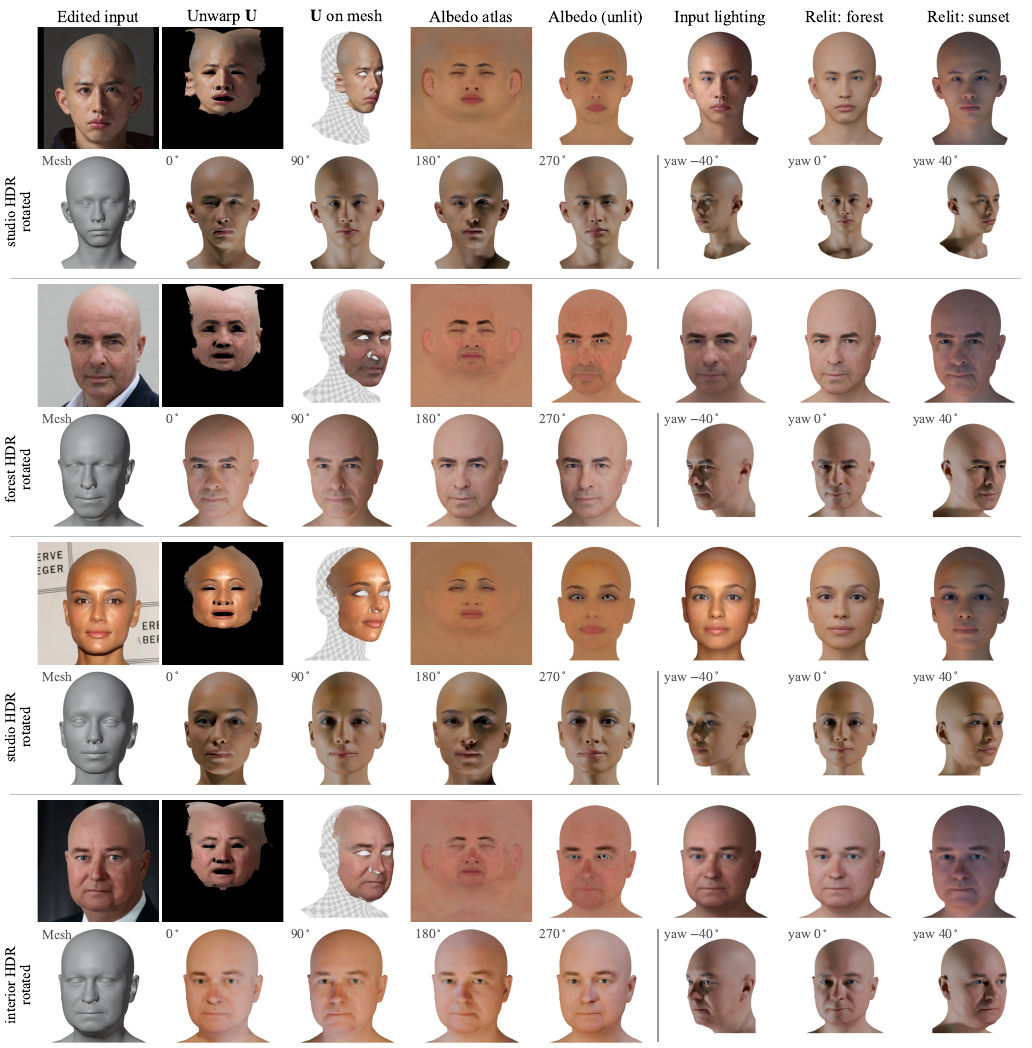}
\caption{\textbf{Texture synthesis on reconstructed meshes.} Two rows per subject. Top, from left: edited input, unwarp $\mathbf U$, $\mathbf U$ applied to the mesh and viewed at yaw $40^\circ$ (checkerboard: texels without image support), albedo atlas completed with the geometry adapter, the albedo rendered without shading, a render under the lighting fitted to the input photograph, and physically based renders under two HDR environments. Bottom: reconstructed mesh; renders with the HDR environment named at left rotated about the vertical axis by $0^\circ$--$270^\circ$, with mesh, albedo, material, and camera fixed; and renders at yaw $-40^\circ$, $0^\circ$, and $40^\circ$ under Blender's studio environment.}
\label{fig:tex_qual}
\end{figure*}

\paragraph{Correspondence construction.}
\label{sec:supp_correspondence}
Each target token is assigned a nominal $16\times16$ texel window, rather than the VAE's exact receptive field.
UV triangles are rasterized at $1024^2$ with pixel centers at $(u\cdot1024-0.5,\,(1-v)\cdot1024-0.5)$.
Let $\Omega_j$ be the nominal texel window of UV token $j$, and let $\kappa(i)$ assign template vertex $i$ to its geometry cluster.
For a texel $p$, $\mathcal F_p$ is the set of triangles covering it and $\beta_i^f(p)$ is its barycentric weight at vertex $i$ of triangle $f$.
The bias is constructed from the accumulated overlap $\mathbf O$ with entries
\begin{equation}
O_{jk}=\sum_{p\in\Omega_j}\frac{1}{|\mathcal F_p|}\sum_{f\in\mathcal F_p}\sum_{i\in f}\beta^{f}_i(p)\,\mathbb{1}[\kappa(i)=k],
\label{eq:overlap}
\end{equation}
where triangles covering the same texel share its unit area equally; uncovered texels ($\mathcal F_p=\emptyset$) contribute zero.
For rows with positive coverage, $W_{jk}=O_{jk}/\sum_{k'}O_{jk'}$; uncovered rows of $\mathbf W$ remain zero.
Unlike a cluster's mean UV coordinate, these weights retain contributions from disconnected UV islands and seams.
Two clusters are adjacent if they share a triangle. With row-normalized adjacency $\hat{\mathbf A}$ including self-loops, the one-hop expansion is $\widetilde{\mathbf W}=\mathrm{norm}(\mathbf W+\tfrac12\mathbf W\hat{\mathbf A})$, where $\mathrm{norm}$ normalizes nonzero rows.
The attention bias $\mathbf B=\log(\widetilde{\mathbf W}+10^{-6})$ is fixed across subjects; the geometry memory $\mathbf G$ changes with each reconstruction.
Only covered target tokens, indexed by $\mathcal J$, receive an adapter residual.
\Cref{tab:supp_corr} reports statistics of $\mathbf W$.

\begin{table}[h]
\centering
\caption{Statistics of the correspondence matrix.}
\label{tab:supp_corr}
\small
\resizebox{\linewidth}{!}{%
\begin{tabular}{l c}
\toprule
Shape & $4{,}096\times1{,}961$ \\
Covered / uncovered tokens & 4{,}036 / 60 \\
Clusters per covered token (min / median / max) & 1 / 4 / 48 \\
\quad with the 1-hop halo & 4 / 14 / 87 \\
Nonzeros (density), without / with halo & 17{,}585 (0.22\%) / 62{,}249 (0.77\%) \\
Clusters without UV coverage & 0 \\
\bottomrule
\end{tabular}}
\end{table}

\paragraph{Adapter computations.}
\label{sec:supp_adapter_computation}
For each selected block $b$, the adapter reads the modulated target-UV hidden states $\tilde{\mathbf Y}^b$ that also enter the frozen block's attention.
Each head projects the covered target states and the shared geometry memory as
\begin{equation}
\begin{aligned}
\mathbf Q_{\mathcal J}&=\RMS(\tilde{\mathbf Y}^b_{\mathcal J}\mathbf P_q^b),\\
\mathbf K_G&=\RMS(\mathbf G\mathbf P_k^b),\qquad \mathbf V_G=\mathbf G\mathbf P_v^b,
\end{aligned}
\label{eq:supp_adapter_projections}
\end{equation}
with per-head RMSNorm on queries and keys; projection biases are omitted for readability.
Reference tokens receive no direct adapter residual.
Attention weights are proportional to $\exp(\mathbf Q_{\mathcal J}\mathbf K_G^\top/\sqrt{d_h})\odot(\widetilde{\mathbf W}_{\mathcal J}+10^{-6})$: learned content scores modulate the fixed spatial preference rather than imposing a hard correspondence mask.
Zero-initializing the output projection's weight and bias makes the added residual vanish, so the adapter starts from the frozen block's function.

\section{Texture synthesis}
\label{sec:supp_tex}
Section~\ref{sec:texture} describes how the texture model completes a partial UV atlas. This section details the FaceUV backbone, camera fit and adapter training. Figure~\ref{fig:tex_qual} shows the same atlas under different lighting conditions and viewpoints.

\paragraph{Base FaceUV model.}
We use Qwen-Image-Edit-2511~\cite{qwenimage} with a fused rank-256 FaceUV LoRA~\cite{lora}, an empty prompt, and a $1024^2$ output atlas.
The transformer has 60 dual-stream blocks of width 3072. Its target UV latent has shape $16\times128\times128$, patchified with $2\times2$ patches into 4{,}096 target tokens.
The main LoRA predicts albedo from lit portraits. A separate LoRA trained for the same two-reference task completes the texture under the input lighting for comparison with real scans (\cref{tab:tex}).

\paragraph{Camera and unwarping.}
We fit a weak-perspective camera to 16 MediaPipe landmarks around the eyes, nose, and mouth and their corresponding vertices on the reconstructed head. To establish these correspondences, we use 300 synthetic renders with known cameras and assign each landmark its most frequent nearest projected vertex, using matches within 15 pixels at $512^2$.
The mask $\mathbf M$ includes both projected texels and locally inpainted gaps~\cite{telea}; it is not a strict visibility mask.

The side view in \cref{fig:tex_qual} shows where the completed atlas fills gaps in the unwarp on the head and neck. Rotating the HDR environment relights the same predicted atlas. The Cycles material uses the predicted albedo, constant roughness, weak subsurface scattering, a shared eye texture, and no normal map.

\paragraph{Adapter configuration.}
The adapter uses eight heads in blocks $\{6,14,21,29,36,44,51,59\}$ (zero-based). It projects queries from 3072 to 1024 channels, keys and values from the 512-channel geometry descriptor $\phi$ (\cref{eq:geomem}) to 1024, and the output back to 3072.
The geometry descriptors are computed in FP32.
The descriptor MLP is LayerNorm, Linear($518\rightarrow512$), SiLU, and Linear($512\rightarrow512$), giving $d_g=512$. Together with the adapter branches it contains 59.3M trainable parameters; the editing backbone, fused LoRA, and graph VAE remain frozen.
The cluster positions $\bar{\bm c}_k$ are obtained by repeatedly mean-pooling vertex coordinates over immediate children with the parent maps of the VAE. Because child clusters have unequal sizes, this hierarchical mean need not equal the arithmetic centroid of all original vertices in a coarse cluster.
Vertex normals are formed by summing incident unnormalized face cross products and normalizing, then pooled in the same way and normalized again to give $\bar{\bm n}_k$.
The adapter is trained with AdamW (lr $10^{-4}$) and effective batch 32.
Sampling uses 10 steps and classifier-free guidance scale 4.

\paragraph{Texture flow schedule and weighting.}
We sample $i$ uniformly from the Qwen scheduler's 1{,}000 entries. With $r_i=1-i/1000$ and $\tau_i=e^{0.8}/(e^{0.8}+1/r_i-1)$, the noise level is $t_i=1-0.98(1-\tau_i)/(1-\tau_{999})$ for $i=0,\ldots,999$, ending at $0.02$.
The model receives timestep $1000t_i$ and noisy latent $(1-t_i)\mathbf y_0+t_i\bm\epsilon$.
The mean-squared velocity loss is multiplied by
\begin{equation}
w_i=\frac{1000(a_i-a_{\min})}{\sum_{j=0}^{999}(a_j-a_{\min})},\quad
a_i=\exp[-2(t_i-\tfrac12)^2],
\end{equation}
where $a_{\min}=\min_j a_j$ and $w(t_i)=w_i$ in \cref{eq:texloss}.

\section*{Experiments and Analysis}
\addcontentsline{toc}{section}{Experiments and Analysis}

\section{Data and evaluation protocols}
\label{sec:supp_data}
\label{sec:supp_baselines}
Sections~\ref{sec:data_construction} and~\ref{sec:setup} introduce the datasets and evaluation metrics. This section describes data sampling and splits, mesh alignment and texture scoring. It also specifies the protocols for the template-bias ablation and the TOPOS comparison.

\paragraph{Regional shape sampling.}
The nine regions in \cref{eq:shape_augmentation} are the brows, eyes, nose, mouth, ears, forehead, cheekbones, face outline, and chin. For the three targets selected per region, we draw weights independently and uniformly over their artist-specified ranges. The median whole-head V2V between two instances of the same base character is $19\times10^{-3}$ without ICP (\cref{fig:supp_datavar}a).
\paragraph{Appearance variation in the atlas.}
The Gaussian filter in \cref{eq:skintone_augmentation} uses a standard deviation of 31 texels, reflected padding, and a $1024^2$ grid. The two skin-tone coefficients specify a global RGB color in a two-component PCA model and a spatial offset in a $128^2$ UV basis; the first coefficient sets the range of the second. We add the color and offset, then bilinearly upsample to obtain $\mathbf L(\bm\beta)$ (\cref{fig:supp_datavar}b).

\paragraph{Portraits and dense correspondence labels.}
For a selected mesh and texture, we render a frontal view (yaw $0^\circ$, pitch $10^\circ$) and sample two further views with yaw in $[-60^\circ,60^\circ]$ and pitch in $[-15^\circ,30^\circ]$.
We render orthographic portraits at $2048^2$ under uniform or sampled HDR lighting and use aligned $512^2$ inputs for CCE.
To produce a CCE target we restore the head and its recorded camera, replace the appearance material with an unlit shader that writes $(u,v,0)$, and save the result as a 16-bit image.
The same coordinate image serves all lighting conditions for a given mesh and camera.

\paragraph{Training splits.}
The geometry adapter follows the CCE and geometry-flow character split. Training and test atlases are also disjoint by source map: no tone, eyebrow, or makeup variant of a test atlas occurs in training.
Only the full geometry flow also uses registered LYHM~\cite{lyhm,lyhm_ijcv2020} and FaceScape~\cite{facescape} scans paired with hair-removed portraits. These subjects are disjoint from the real-scan test set.

\paragraph{Cross-topology geometry evaluation.}
\begin{figure*}[!tp]
\centering
\includegraphics[width=\textwidth]{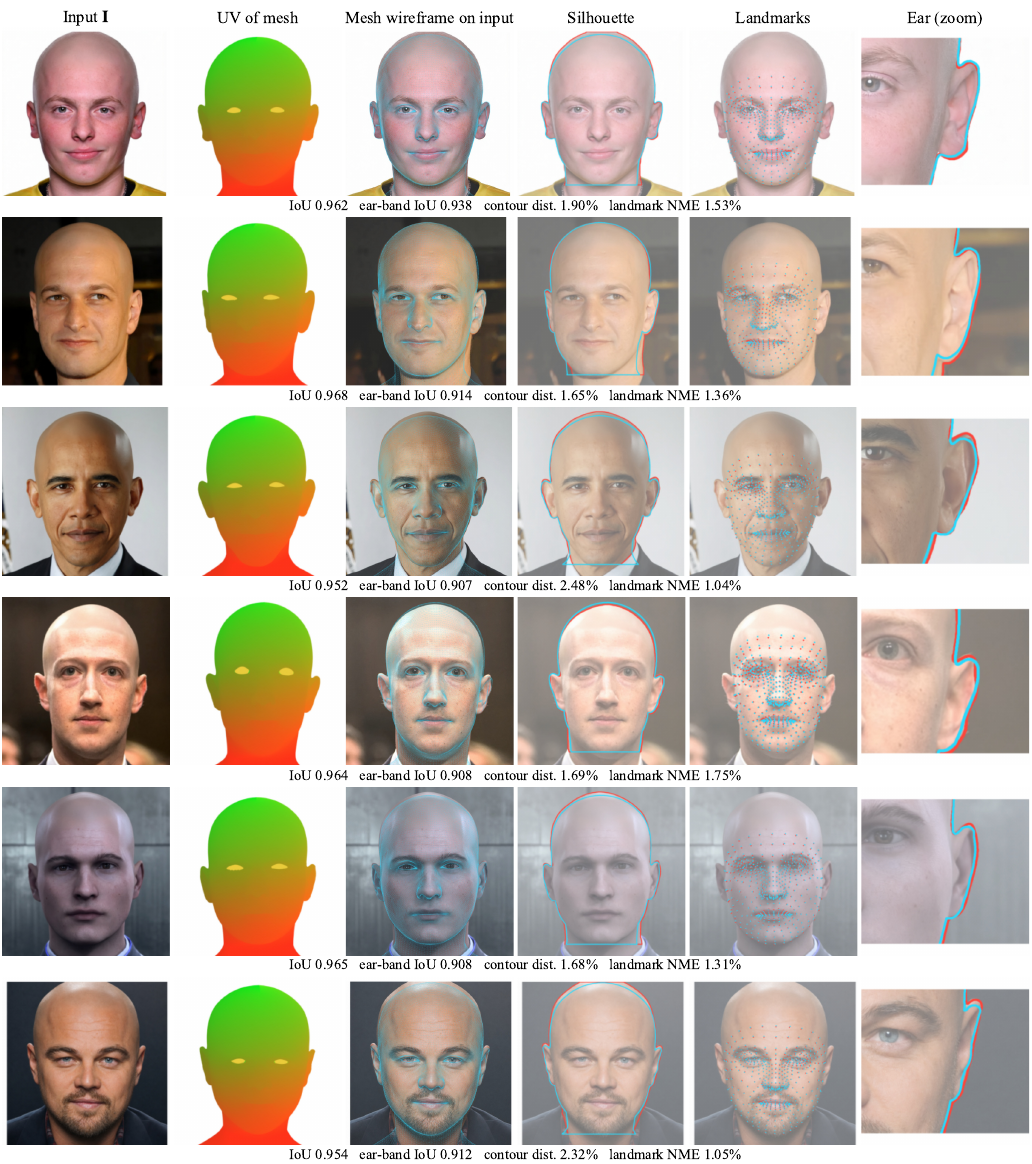}\par\smallskip
\begin{minipage}[t]{0.69\textwidth}
\vspace{0pt}
\caption{\textbf{2D alignment with the edited portrait.} Left to right: edited input, projected UV coordinates, wireframe overlay, head contours, facial landmarks, and an ear detail. For contours and landmarks, red denotes the portrait and cyan the reconstruction. Rows show individual cases; the summary covers 59 in-the-wild portraits. Distances are in percent of inter-ocular distance (IOD).}
\label{fig:supp_align}
\end{minipage}\hfill
\begin{minipage}[t]{0.29\textwidth}
\vspace{0pt}
\centering\footnotesize\setlength{\tabcolsep}{2pt}
\begin{tabular}{@{}lrr@{}}
\toprule
Metric & Mean & Median \\
\midrule
Silhouette IoU $\uparrow$ & 0.948 & 0.949 \\
Ear-band IoU $\uparrow$ & 0.845 & 0.846 \\
Contour dist.\ $\downarrow$ & 2.81 & 2.78 \\
Landmark NME $\downarrow$ & 1.31 & 1.27 \\
\bottomrule
\end{tabular}
\end{minipage}
\end{figure*}
For \cref{tab:geo,tab:supp_geo_full}, we use each baseline's released weights.
DECA contributes its detail mesh and MICA its neutral identity mesh; Hunyuan3D~2.1 and TRELLIS.2 run their shape-generation pipelines, the latter with its 1024-resolution cascade, and Pixal3D runs zero-shot at resolution 1024 with a fixed approximate camera.
The seven landmarks used for alignment are the four eye corners, nose base, and two mouth corners. We use fixed annotations for the ground truth and template-based outputs, and MediaPipe~\cite{mediapipe} detections on a grey render for other topologies. ICP uses 4{,}096 ground-truth samples, at most 1{,}000 iterations, and a relative tolerance of $10^{-5}$.
For scoring, we sample 20{,}000 points uniformly by area on the ground-truth face and find the nearest point on the full, uncropped prediction. Distances use the outer-eye-corner normalization of \cref{sec:setup}.
The facial region is defined per dataset from the ground truth: a landmark-derived region from the brows to the chin with the eye and mouth openings removed for LYHM and FaceScape, and the face region of the template's exterior skin for the synthetic set; scores are therefore not pooled across datasets.
Because the distance is one-directional, surface outside the ground-truth region is not penalized, and the metric applies to any topology.
A case is excluded if it has an invalid ground-truth landmark annotation, a failed landmark detection, or an alignment that did not converge.
FaceScape and synthetic results are averaged over cases valid for all methods; LYHM results are averaged over each method's valid cases.
For \cref{fig:geo_qual}, the aligned meshes of all methods are rendered with the camera of the input portrait: the released calibration for FaceScape, a camera estimated by matching the scan texture to the portrait for LYHM, and the rendering camera for the synthetic set.

\renewcommand{\floatpagefraction}{0.90}
\raggedbottom
\begin{table}[!htbp]
\centering
\caption{\textbf{Facial-region geometry.} $^\dagger$TOPOS is added to \cref{tab:geo} under the same protocol. Distances and recall thresholds are in \% OCD; recall is in \%. Bold: best; underline: second.}
\label{tab:supp_geo_full}
\scriptsize
\renewcommand{\arraystretch}{0.90}
\setlength{\tabcolsep}{1.25pt}
\begin{tabular}{@{}l ccc c ccc@{}}
\toprule
Method & S2M$\downarrow$ & Median$\downarrow$ & RMSE$\downarrow$ & NC$\uparrow$ & R@0.5\%$\uparrow$ & R@1\%$\uparrow$ & R@2\%$\uparrow$ \\
\midrule
\multicolumn{8}{@{}l}{\emph{LYHM~\cite{lyhm}}} \\
DECA~\cite{deca} & 1.282 & 1.068 & 1.626 & 0.950 & 25.9 & 48.8 & 79.3 \\
MICA~\cite{mica} & \second{0.986} & \second{0.816} & \best{1.258} & \second{0.956} & 33.4 & 60.4 & \best{89.1} \\
SMIRK~\cite{smirk} & 1.280 & 1.054 & 1.633 & 0.947 & 26.3 & 49.2 & 79.3 \\
HRN~\cite{hrn} & 1.221 & 1.009 & 1.560 & 0.949 & 27.1 & 50.8 & 81.4 \\
Pixel3DMM~\cite{pixel3dmm} & 1.224 & 1.025 & 1.549 & 0.951 & 27.2 & 50.8 & 81.3 \\
Hunyuan3D 2.1~\cite{hunyuan3d21} & 1.097 & 0.827 & 1.466 & 0.954 & \best{35.5} & \second{60.6} & 83.8 \\
TRELLIS.2~\cite{trellis2} & 1.267 & 1.037 & 1.625 & 0.942 & 27.0 & 49.9 & 79.5 \\
Pixal3D~\cite{pixal3d} & 1.371 & 1.125 & 1.762 & 0.921 & 24.6 & 46.5 & 76.7 \\
TOPOS$^\dagger$~\cite{topos} & 1.462 & 1.170 & 1.890 & 0.938 & 24.0 & 45.0 & 74.3 \\
\method (ours) & \best{0.978} & \best{0.798} & \best{1.258} & \best{0.962} & \second{34.2} & \best{61.3} & \second{88.8} \\
\midrule
\multicolumn{8}{@{}l}{\emph{FaceScape~\cite{facescape}}} \\
DECA~\cite{deca} & 1.660 & 1.377 & 2.109 & 0.930 & 20.3 & 38.7 & 67.4 \\
MICA~\cite{mica} & 0.985 & 0.814 & 1.254 & 0.951 & 33.3 & 60.6 & 89.0 \\
SMIRK~\cite{smirk} & 1.404 & 1.165 & 1.793 & 0.930 & 23.9 & 45.0 & 75.4 \\
HRN~\cite{hrn} & 1.322 & 1.117 & 1.659 & 0.948 & 24.3 & 46.3 & 77.6 \\
Pixel3DMM~\cite{pixel3dmm} & \best{0.867} & \best{0.722} & \best{1.105} & \second{0.954} & \best{37.6} & \best{66.1} & \best{92.5} \\
Hunyuan3D 2.1~\cite{hunyuan3d21} & 1.347 & 1.005 & 1.799 & 0.936 & 28.4 & 54.0 & 76.9 \\
TRELLIS.2~\cite{trellis2} & 1.310 & 1.093 & 1.663 & 0.939 & 26.0 & 48.9 & 78.9 \\
Pixal3D~\cite{pixal3d} & 1.211 & 1.012 & 1.539 & 0.927 & 26.9 & 50.6 & 81.6 \\
TOPOS$^\dagger$~\cite{topos} & 1.393 & 1.173 & 1.766 & 0.939 & 23.6 & 44.7 & 75.7 \\
\method (ours) & \second{0.892} & \second{0.753} & \second{1.129} & \best{0.964} & \second{35.8} & \second{64.3} & \second{92.3} \\
\midrule
\multicolumn{8}{@{}l}{\emph{Synthetic}} \\
DECA~\cite{deca} & 1.896 & 1.565 & 2.423 & 0.932 & 18.4 & 35.5 & 62.6 \\
MICA~\cite{mica} & 1.763 & 1.457 & 2.249 & 0.934 & 19.3 & 37.3 & 65.3 \\
SMIRK~\cite{smirk} & 1.903 & 1.575 & 2.419 & 0.932 & 18.2 & 35.1 & 62.4 \\
HRN~\cite{hrn} & 2.446 & 1.984 & 3.312 & 0.931 & 13.9 & 27.4 & 51.5 \\
Pixel3DMM~\cite{pixel3dmm} & 1.698 & 1.410 & 2.165 & 0.933 & 20.4 & 38.9 & 67.3 \\
Hunyuan3D 2.1~\cite{hunyuan3d21} & \second{1.275} & \best{0.911} & \second{1.769} & 0.930 & \best{29.6} & \best{57.1} & \second{80.3} \\
TRELLIS.2~\cite{trellis2} & 1.970 & 1.582 & 2.550 & 0.914 & 17.9 & 34.3 & 60.9 \\
Pixal3D~\cite{pixal3d} & 1.899 & 1.542 & 2.463 & 0.911 & 18.7 & 35.7 & 62.6 \\
TOPOS$^\dagger$~\cite{topos} & 1.492 & 1.222 & 1.928 & \second{0.947} & 24.0 & 45.1 & 74.0 \\
\method (ours) & \best{1.197} & \second{0.973} & \best{1.545} & \best{0.960} & \second{28.7} & \second{53.0} & \best{82.2} \\
\bottomrule
\end{tabular}
\end{table}

\newcommand{\suppToposWildFigure}{%
\begin{figure*}[!t]
\centering
\includegraphics[width=\textwidth]{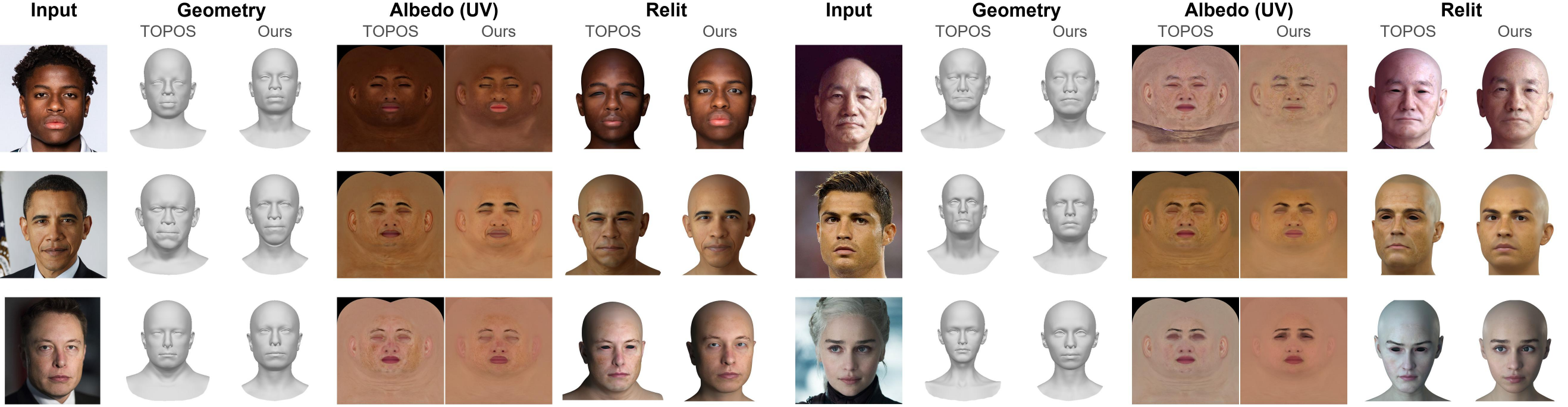}
\caption{\textbf{TOPOS~\cite{topos} on real portraits.} Both methods use the same edited bald input. Each photograph is followed by paired geometry, UV albedo and renders (TOPOS $|$ Ours). We render each atlas on its own mesh under lighting estimated by \method; TOPOS renders omit eyeballs.}
\label{fig:supp_topos_wild}
\end{figure*}}

\paragraph{Whole-head geometry.}
\label{sec:supp_geo_head}
For \cref{tab:supp_geo_head}, we keep the alignment and metrics of \cref{tab:geo} but sample the entire ground-truth head, including the scalp, ears, and neck. Each sampled point is compared with the nearest point on the predicted surface.

\begin{table}[!htbp]
\centering
\caption{\textbf{Whole-head geometry on the synthetic test set.} Face reports facial S2M from \cref{tab:geo}; Head evaluates the full head, including the scalp, ears, and neck. S2M in percent of OCD. Bold: best.}
\label{tab:supp_geo_head}
\normalsize
\setlength{\tabcolsep}{5pt}
\begin{tabular}{@{}l c ccc@{}}
\toprule
& Face & \multicolumn{3}{c@{}}{Head} \\
\cmidrule(lr){2-2}\cmidrule(l){3-5}
Method & S2M$\downarrow$ & S2M$\downarrow$ & NC$\uparrow$ & R@1\%$\uparrow$ \\
\midrule
DECA~\cite{deca} & 1.896 & 6.926 & 0.942 & 15.38 \\
MICA~\cite{mica} & 1.763 & 5.797 & 0.944 & 18.00 \\
SMIRK~\cite{smirk} & 1.903 & 6.106 & 0.943 & 17.03 \\
HRN~\cite{hrn} & 2.446 & 58.224 & 0.694 & 7.27 \\
Pixel3DMM~\cite{pixel3dmm} & 1.698 & 5.398 & 0.944 & 19.14 \\
Hunyuan3D 2.1~\cite{hunyuan3d21} & 1.275 & 4.756 & 0.943 & \best{32.01} \\
TRELLIS.2~\cite{trellis2} & 1.970 & 6.047 & 0.938 & 17.43 \\
Pixal3D~\cite{pixal3d} & 1.899 & 6.781 & 0.927 & 17.59 \\
\midrule
\method (ours) & \best{1.197} & \best{4.260} & \best{0.965} & 27.44 \\
\bottomrule
\end{tabular}
\end{table}

\paragraph{2D alignment with the edited portrait.}
In \cref{fig:supp_align}, we compare each edited portrait with its textured render under the estimated camera and fitted lights, using matching crops with a detected face. Above the chin, we measure silhouette IoU and symmetric contour distance against the portrait's head mask (MediaPipe hair and skin classes); ear-band IoU is restricted to the region outside the face oval between the brows and nose base. Landmark NME compares the 478 MediaPipe landmarks detected on the portrait and on the render.

\newcommand{\suppBiasTableContents}{%
\centering
\caption{\textbf{Template attention bias.} Unlit albedo on 200 synthetic subjects. No-adapter and full-model scores are from \cref{tab:tex}.}
\label{tab:supp_adapter_bias}
\normalsize
\setlength{\tabcolsep}{4pt}
\begin{tabular}{@{}l ccc@{}}
\toprule
Variant & PSNR$\uparrow$ & SSIM$\uparrow$ & LPIPS$\downarrow$ \\
\midrule
No adapter & 24.13 & 0.869 & 0.153 \\
Adapter, $\mathbf B=0$ & 24.18 & 0.871 & 0.147 \\
Full adapter & \best{24.67} & \best{0.876} & \best{0.125} \\
\bottomrule
\end{tabular}}

\newcommand{\suppTextureAblations}{%
\paragraph{Texture ablations.}
Each column of \cref{tab:tex} uses one LoRA and the same per-case noise seeds. The no-unwarp variant omits $\mathbf U$ in training and inference. The $\mathbf B=0$ control disables only the bias in both stages, retaining geometry memory, trainable projections and covered UV tokens. Restoring the bias raises PSNR by 0.49~dB and SSIM by 0.005 and lowers LPIPS by 0.022 (\cref{tab:supp_adapter_bias}).\par}
\newcommand{\suppTextureAblationBlock}{%
\begin{samepage}
\suppTextureAblations
\nopagebreak\medskip\noindent
\begin{minipage}{\columnwidth}
\edef\suppPreviousTable{\number\value{table}}
\setcounter{table}{10}
\captionsetup{type=table}
\suppBiasTableContents
\setcounter{table}{\suppPreviousTable}
\end{minipage}\par
\end{samepage}}

\paragraph{Texture evaluation.}
\emph{Real scans: appearance under capture lighting.} We use the baked-lighting LoRA for the LYHM and released FaceScape TU textures; MARCUS uses its inpainting output. \emph{Synthetic: unlit albedo renders.} FFHQ-UV, MARCUS, and \method use their albedo outputs; each albedo atlas is mapped onto the method's reconstructed mesh and rendered without lighting, and compared with the same unlit render of the ground-truth albedo. Both comparisons score each method's own textured mesh~\cite{uvidm}, not an atlas in a common UV layout.

\newcommand{\suppToposTable}{%
\centering
\captionsetup{type=table}
\caption{\textbf{Unlit albedo on 200 synthetic subjects.} Each method uses its own mesh (\cref{tab:tex}); PSNR is in dB.}
\label{tab:supp_topos}
\normalsize
\setlength{\tabcolsep}{4pt}
\begin{tabular}{@{}l ccc@{}}
\toprule
Method & PSNR$\uparrow$ & SSIM$\uparrow$ & LPIPS$\downarrow$ \\
\midrule
TOPOS~\cite{topos} & 23.13 & 0.841 & 0.192 \\
\method (ours) & \best{24.67} & \best{0.876} & \best{0.125} \\
\bottomrule
\end{tabular}}

\suppToposWildFigure

\emph{Rendering.} The similarity alignment from geometry evaluation is used only for scoring. We render prediction and reference at $512^2$ at yaw $0^\circ$ and $\pm30^\circ$, with linear-light color interpolation and no shading, normal maps, color correction, or exposure normalization.
We use native outputs: $1024^2$ atlases for FFHQ-UV, FreeUV (FLAME), MARCUS (homogenized HiFi3D++), and \method; HRN's $256^2$ mid-level mesh texture; and UV-IDM's 35{,}709 vertex colors.

\emph{Scoring.} The ground-truth face mask excludes eye and mouth openings. Uncovered pixels count as errors.
We compute PSNR in sRGB, SSIM~\cite{ssim} with an $11\times11$ Gaussian window ($\sigma=1.5$) on the mask eroded by 5 pixels, and AlexNet LPIPS~\cite{lpips} by averaging its spatial map over the mask with a common neutral color outside. For all methods, we average the three views per subject before averaging over all 300 real subjects or all 200 synthetic subjects.

\newcommand{\suppToposComparison}{%
\paragraph{Comparison with TOPOS.}
\label{sec:supp_topos}
Both methods use the MetaHuman topology and UV layout. \method has lower mean facial surface error and higher normal consistency on all three datasets (\cref{tab:supp_geo_full}). On synthetic albedo it gains 1.54~dB over TOPOS, with higher SSIM and lower LPIPS (\cref{tab:supp_topos}).\par}

\par\medskip\noindent\begin{minipage}{\columnwidth}
\suppToposComparison\smallskip
\suppToposTable
\end{minipage}\par
\paragraph{In-the-wild comparison.}
In \cref{fig:supp_topos_wild}, TOPOS retains highlights and shadows from the input: relighting darkens the eye sockets and nasolabial folds a second time (middle row). Skin tone follows the illuminant in the top- and bottom-right examples. The collar in the top-right atlas reflects its reliance on the photograph; misplaced eye features and dark atlas borders suggest misalignment. Side and low-angle views also reveal shape distortion. Our atlas can be relit without baking in the input lighting (\cref{fig:tex_qual}).

\suppTextureAblationBlock

\newcommand{\suppEditingPlate}{%
\centering
\captionsetup{skip=2pt,hypcap=false}
\includegraphics[width=\linewidth]{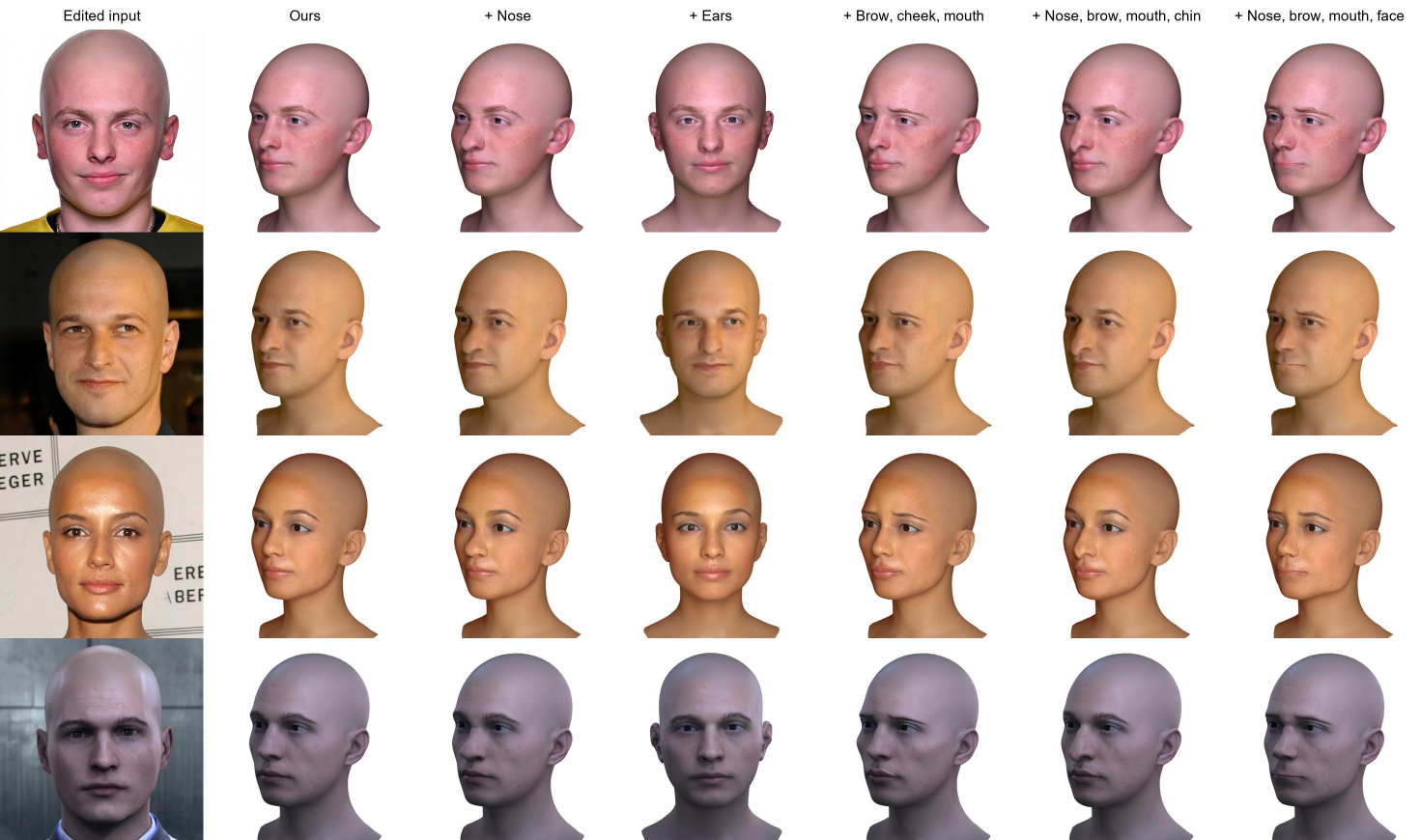}
\captionof{figure}{\textbf{Identity blendshapes on reconstructions.} Ear edits are shown from the front, other edits in three-quarter view. The last two columns combine targets. All views use the reconstructed atlas.}
\label{fig:supp_bs}
\par\smallskip
\includegraphics[width=\linewidth]{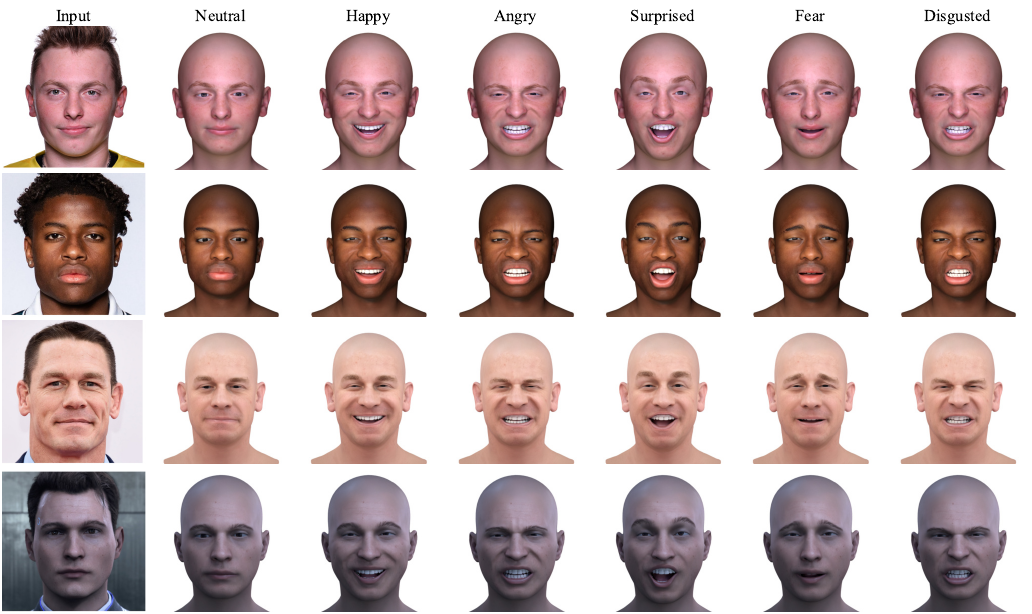}
\captionof{figure}{\textbf{Automatic rig binding.} Each row shows an input, its neutral reconstruction and a frame from each of five expression clips. The same MetaHuman controls drive the head, teeth and eyes across subjects without changing the texture.}
\label{fig:supp_rig}}
\newpage
\section{Results and applications}
\label{sec:supp_results}
Section~\ref{sec:exp_geo} evaluates facial geometry. Here we report whole-head errors, show template edits and assess hair removal on real portraits.

\paragraph{Full geometry metrics.}
\label{sec:supp_geo_full}
On synthetic faces, Hunyuan3D~2.1 leads on median distance and recall at 0.5\% OCD; \method leads on RMSE and recall at 2\% OCD (\cref{tab:supp_geo_full}). Over the whole head, \method has the lowest S2M and highest normal consistency in \cref{tab:supp_geo_head}.

\paragraph{Template blendshapes and rig.}
After similarity alignment to MetaHuman~\cite{metahuman}, we transfer blendshape offsets by vertex index without new correspondences or texture baking (\cref{fig:supp_bs}). We bind the template rig automatically (\cref{fig:supp_rig}).

\subsection*{Real photographs}
\label{sec:supp_wild_results}

\paragraph{In-the-wild preprocessing.}
\label{sec:supp_wild}
We crop the MediaPipe face oval to $512^2$ (1.4$\times$ expansion, 0.15 vertical shift, black padding). Qwen-Image-Edit~\cite{qwenimage} with a separate LoRA (40 steps, guidance 7.5) removes hair, facial hair and accessories and sets the background to white. Sapiens2~\cite{sapiens2} masks the foreground; the unwarp excludes mouth and teeth. We keep edits whose mean facial color error against the original is below a fixed threshold. \Cref{fig:wild} pairs originals and edits with their reconstructions.

\paragraph{Facial fidelity before and after hair removal.}
Before reconstruction, we compare 295 original--edited pairs: 95 in the wild and 100 each from FaceScape and LYHM. In the original face mask, without re-alignment, we compute PSNR, SSIM, LPIPS and the fraction of pixels with CIEDE2000 $\Delta E_{00}\le3$. CSIM compares ArcFace~\cite{arcface} embeddings by cosine similarity on the whole aligned crop (\cref{tab:supp_id}).

\par\smallskip\noindent\begin{minipage}{\linewidth}
\captionsetup{type=table}
\centering
\caption{\textbf{Facial fidelity after hair removal} on 295 pairs.}
\label{tab:supp_id}
\small
\setlength{\tabcolsep}{4pt}
\begin{tabular}{@{}ccccc@{}}
\toprule
PSNR$\uparrow$ & SSIM$\uparrow$ & LPIPS$\downarrow$ & CSIM$\uparrow$ & $\Delta E_{00}{\le}3$ (\%)$\uparrow$ \\
\midrule
28.24 & 0.911 & 0.065 & 0.851 & 82.1 \\
\bottomrule
\end{tabular}
\end{minipage}\par

\stepcounter{table}%

\onecolumn
\enlargethispage{8bp}
\vspace*{-34bp}
\noindent\begin{minipage}{\textwidth}
\suppEditingPlate
\end{minipage}\par\medskip
\phantomsection
\addcontentsline{toc}{section}{Broader impact}
\label{sec:supp_impact}
\noindent\textbf{Broader impact.} The editing tools in Section~\ref{sec:conclusion} can be used for impersonation. Reconstructing a likeness requires consent and applicable data, model and asset permissions. The framework neither marks assets as synthetic nor records their provenance.

\twocolumn
\newcommand{\suppgalleryplate}[2]{%
  \begin{figure*}[p]
  \vspace*{-23bp}
  \noindent\makebox[\textwidth][c]{%
    \begin{minipage}{\dimexpr\paperwidth-24bp\relax}
    \includegraphics[width=\linewidth]{fig/NERVFace_supp_wild_plate_#1.pdf}\par\nointerlineskip
    \vspace{4bp}\nointerlineskip
    \begin{minipage}[t][28bp][t]{\linewidth}#2\end{minipage}
    \end{minipage}}\par
  \vspace*{-23bp}
  \end{figure*}}
\suppgalleryplate{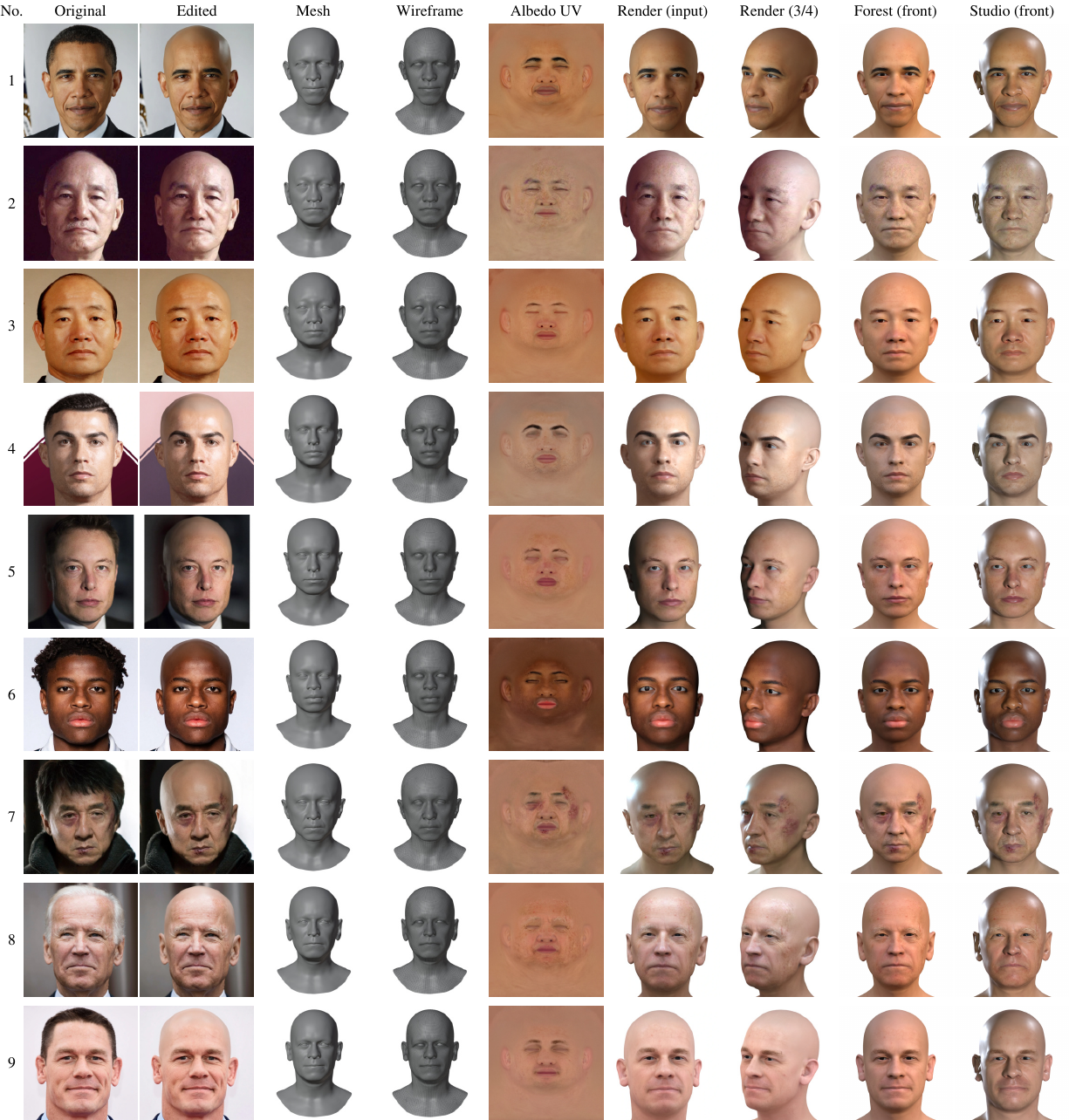}{%
  \caption{\textbf{In-the-wild reconstructions} (portraits 1--9 of 42), extending Figure~\ref{fig:outputs} in Section~\ref{sec:exp_geo}. Input and three-quarter views use lighting fitted to the photograph. The last two columns show the same frontal view under Forest and Studio HDR lighting. Frontal and $35^\circ$ yaw views use $10^\circ$ elevation and the input camera's focal length.}%
  \label{fig:wild}}
\newcommand{\suppgallerypage}[2]{%
  \suppgalleryplate{#1}{\caption{\textbf{In-the-wild reconstructions} (continued; portraits #2 of 42).}}}
\suppgallerypage{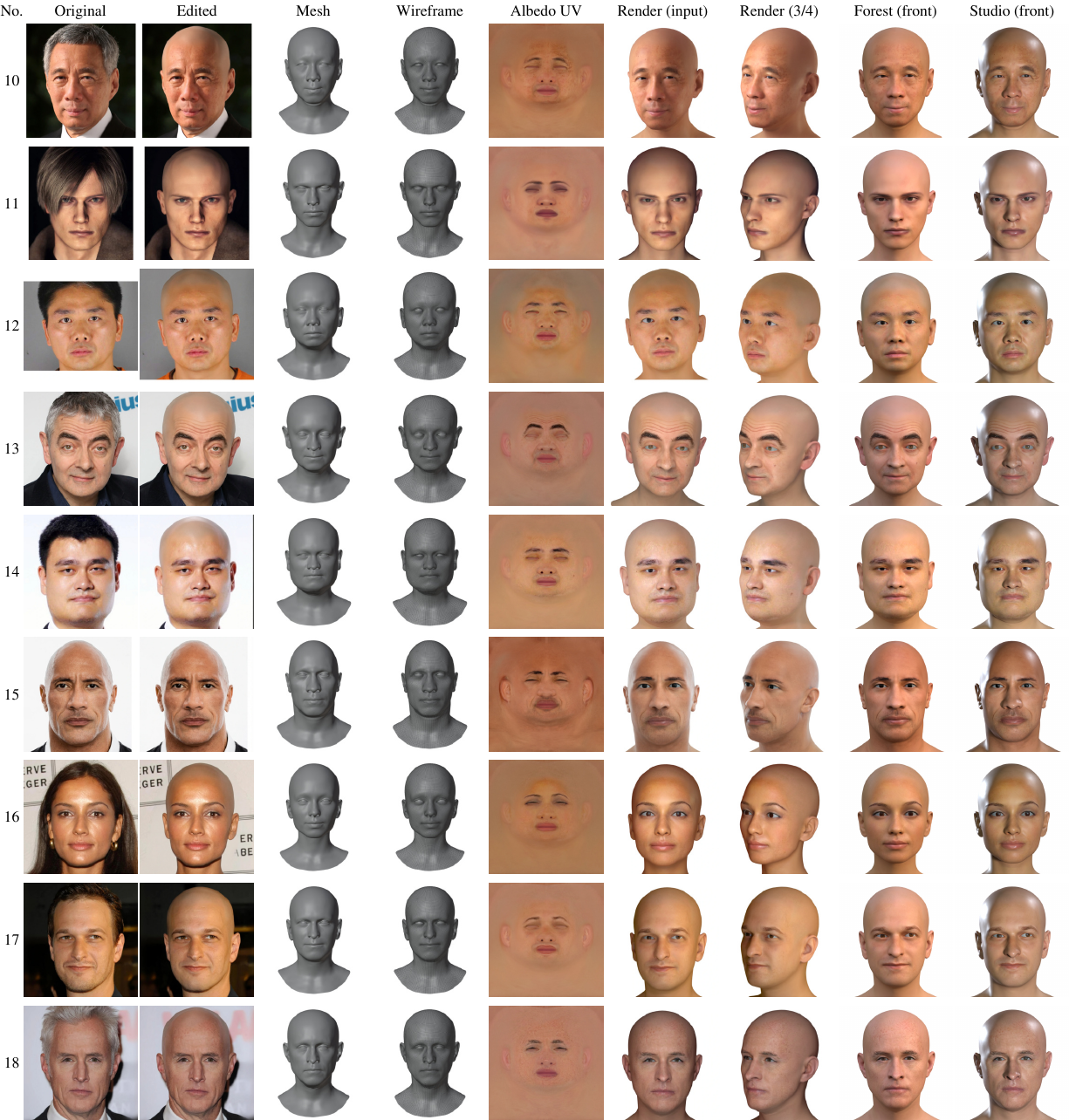}{10--18}
\suppgallerypage{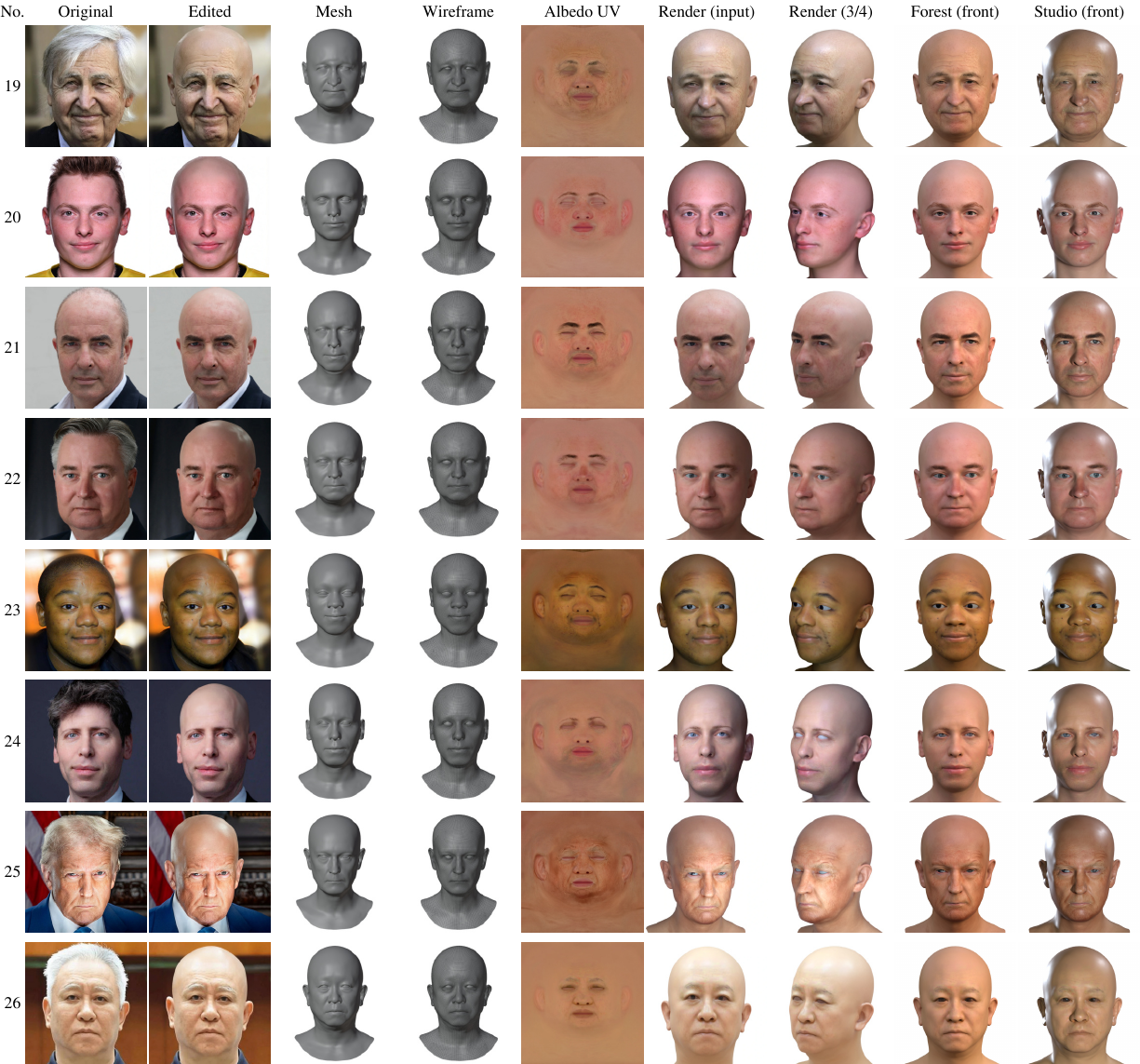}{19--26}
\suppgallerypage{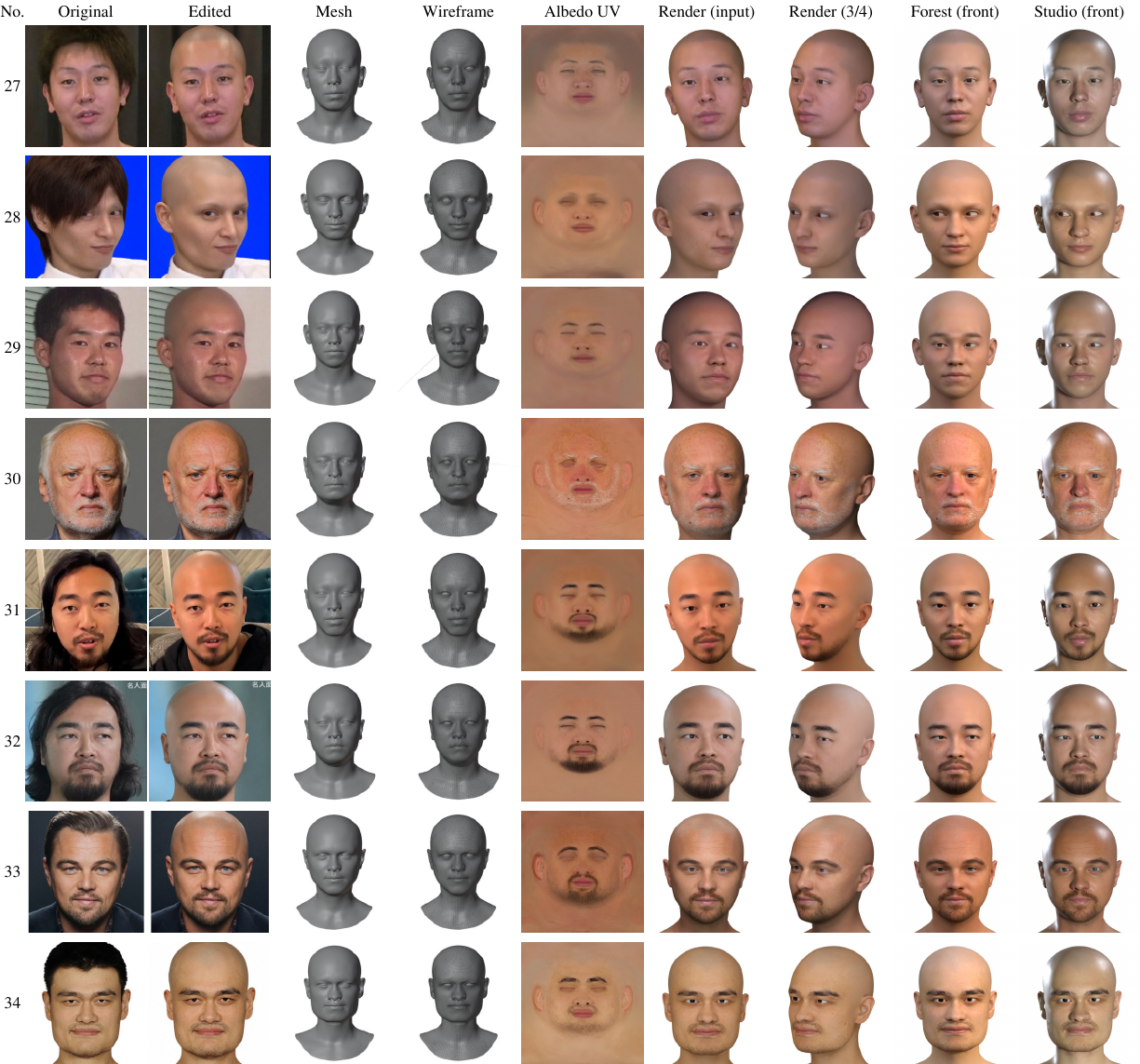}{27--34}
\suppgallerypage{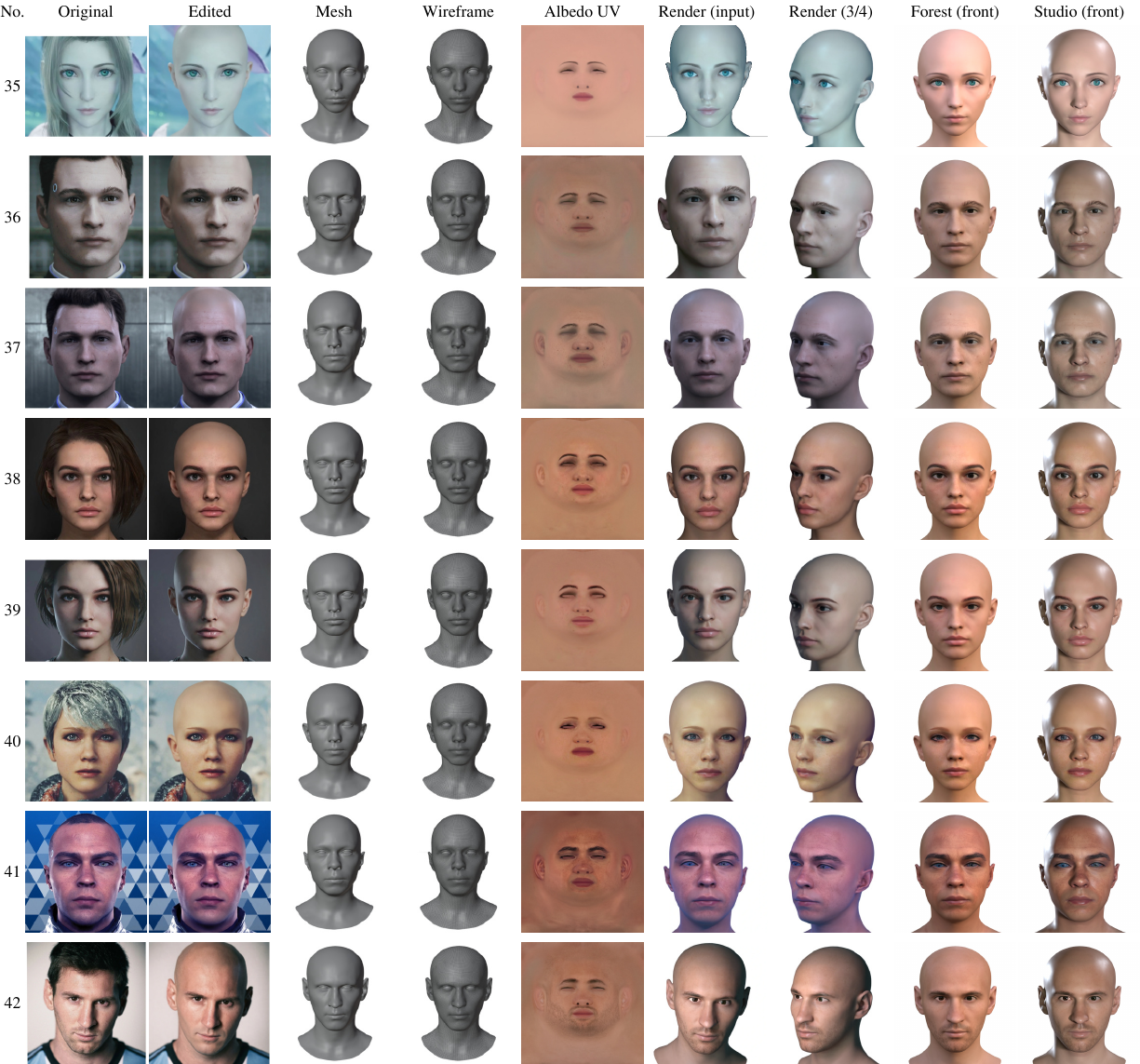}{35--42}

\nocite{ffhquv,freeuv,marcus,omgtex,realy,wang2026industry,paint3d}